\documentclass[runningheads]{llncs}

\newif\ifsupplementary

\usepackage{eccv}

\usepackage{eccvabbrv}

\usepackage{graphicx}
\usepackage{booktabs}

\usepackage[accsupp]{axessibility}  

\usepackage{breqn}
\usepackage{tabularx}

\usepackage[pagebackref,breaklinks,colorlinks,citecolor=eccvblue]{hyperref}

\usepackage{orcidlink}

\begin{document}

\ifsupplementary
\title{Supplementary Material for ``Photorealistic Novel View Synthesis of Human Faces using Next-Scale Transformers''}
\else
\title{Photorealistic Novel View Synthesis of Human Faces using Next-Scale Transformers} 
\fi

\titlerunning{Photorealistic Novel View Synthesis}

\author{Federico Stella\inst{1}\thanks{This work has been carried out during an internship at Meta Reality Labs.} \and
Fei Jiang\inst{2} \and
Zhongshi Jiang\inst{2} \and
Zohar Barzelay\inst{2} \and
Emanuel Garbin\inst{2} \and
Amin Jourabloo\inst{2} \and
Liuhao Ge\inst{2}
}

\authorrunning{F.~Author et al.}

\institute{École polytechnique fédérale de Lausanne, Switzerland \\ \email{federico.stella@epfl.ch} \and
Meta}

\maketitle

\ifsupplementary
  \renewcommand\thefigure{A.\arabic{figure}} 
  \renewcommand\thetable{A.\arabic{table}} 
  \renewcommand\thesection{A.\arabic{section}}
  \setcounter{figure}{0}
  \setcounter{table}{0}
  \setcounter{section}{0}
  \section{Training details}
Our model is trained for 50 hours on 64 NVIDIA A100 GPUs in the first stage. Subsequently, the single-input model is fine-tuned (second stage) for 18 hours on the same hardware. We refer to this model as "Ours". It is specialized for multi-output generation (third stage) for 2 hours. We refer to this model as \textit{Ours MO}. The multi-input model is fine-tuned (second stage) for 14 hours on 128 A100 GPUs, and specialized for multi-output generation (third stage) for 2 hours on the same hardware. We refer to this model as \textit{Ours MI-MO}. The VQ-VAE is fine-tuned on the target dataset for 30 epochs on 8 A100 GPUs, has a codebook size $V=4096$ and uses the following scales for the $K=10$ latent feature maps: (1, 2, 3, 4, 6, 9, 13, 18, 24, 32) for 512x512 images, and (1, 2, 3, 4, 5, 6, 8, 10, 13, 16) for 256x256 images. In the experiments with $K=12$ latent feature maps, we use the following scales: (1, 2, 3, 4, 5, 6, 8, 10, 13, 16, 24, 32).\\

\section{Multiple-output SOS token embedding}
We justify the choice of our multi-view SOS tokens by testing three scenarios using a d12 model at 256x256: a separate SOS token per output view (our model); a shared SOS token across all views, obtained by embedding the concatenation of all the target views; a shared SOS token across all views, obtained without any view information (since the views are always fixed and in a known order). These three approaches achieve a PSNR of 19.99, 19.85 and 19.85 respectively, showing little to no impact. We opted for the first approach since it is the most intuitive and straightforward.

\section{A different scenario: human bodies}
Our method is not limited to human faces, and can be applied to other categories of data as well, such as human bodies. We train our single-input pipeline on the PSGS~\cite{psgs} dataset for human body reconstruction, following the same training process described in the main paper for human faces (including VQ-VAE fine-tuning). We report qualitative results from the validation set in \cref{fig:body_nvs}, showing the canonical views produced by our architecture, and in \cref{fig:body_turntable}, showing the final views rendered after the GS-LRM 3D lifting process. Notice that our single-input single-output model (\textit{Ours}) is not trained to produce only the 6 canonical views, and thus can be used to render any target view of the subject. In the second row of \cref{fig:body_turntable} we show turntable views rendered directly by \textit{Ours} without the lifting step. This demonstrates that our method can be applied to this domain without any architectural changes, showing the flexibility of the method and a potential direction for future work.

\begin{figure*}[ht]
    \centering
    \begin{minipage}[t]{0.48\textwidth}
        \begin{minipage}[c]{0.18\linewidth}
            \centering
            \scriptsize Inputs
        \end{minipage}%
        \hspace{0.01\linewidth}
        \begin{minipage}[c]{0.88\linewidth}
            \includegraphics[width=0.25\linewidth]{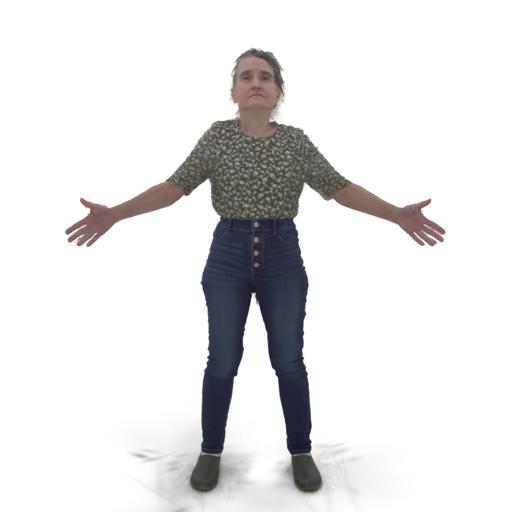}
        \end{minipage}

        \par\nointerlineskip

                \begin{minipage}[c]{0.18\linewidth}
            \centering
            \scriptsize GT
    \end{minipage}%
    \hspace{0.01\linewidth}
    \begin{minipage}[c]{0.8\linewidth}
        \centering
        \includegraphics[width=1\linewidth]{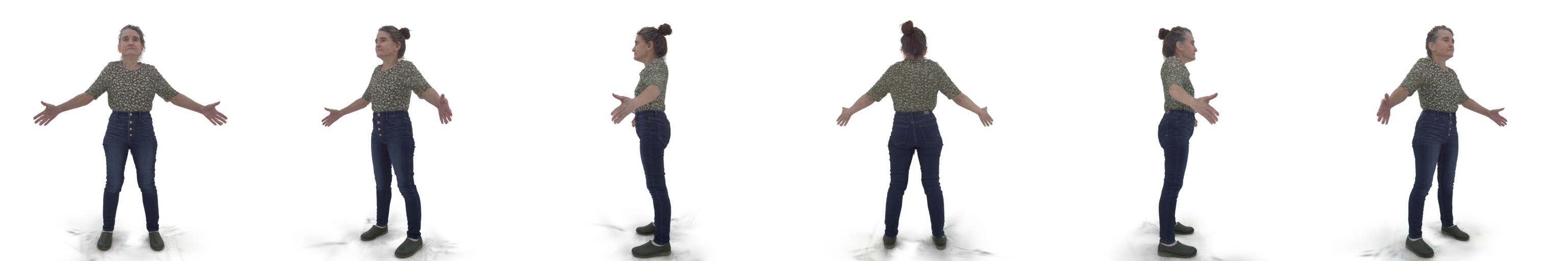}
    \end{minipage}
    \par\nointerlineskip
    \begin{minipage}[c]{0.18\linewidth}
        \centering
        \scriptsize \textit{Ours}
    \end{minipage}%
    \hspace{0.01\linewidth}
    \begin{minipage}[c]{0.8\linewidth}
        \centering
        \includegraphics[width=1\linewidth]{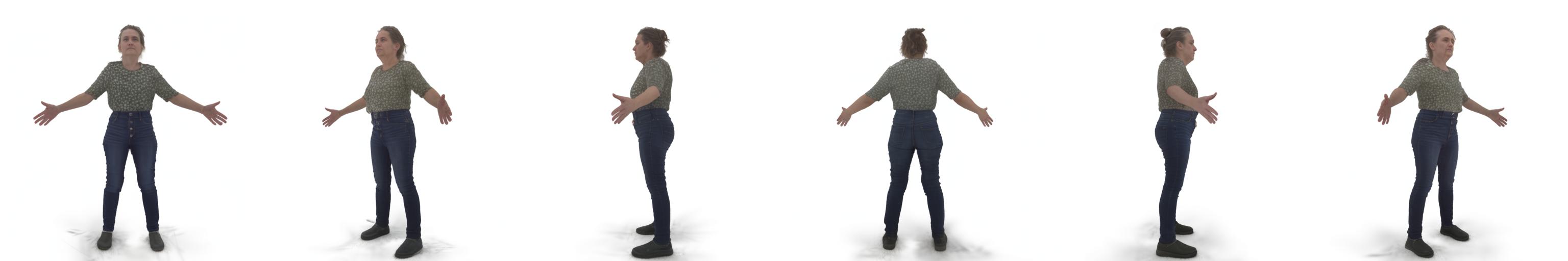}
    \end{minipage}
    \par\nointerlineskip
    \begin{minipage}[c]{0.18\linewidth}
        \centering
        \scriptsize \textit{Ours MO}
    \end{minipage}%
    \hspace{0.01\linewidth}
    \begin{minipage}[c]{0.8\linewidth}
        \centering
        \includegraphics[width=1\linewidth]{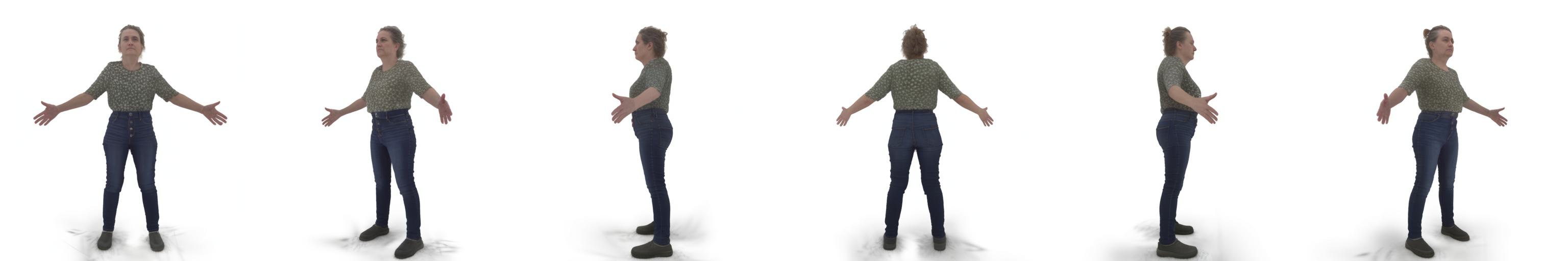}
    \end{minipage}

    \end{minipage}%
    \hspace{0.02\textwidth}
    \begin{minipage}[t]{0.48\textwidth}
        \begin{minipage}[c]{0.88\linewidth}
            \hspace{0.09\linewidth}
            \includegraphics[width=0.25\linewidth]{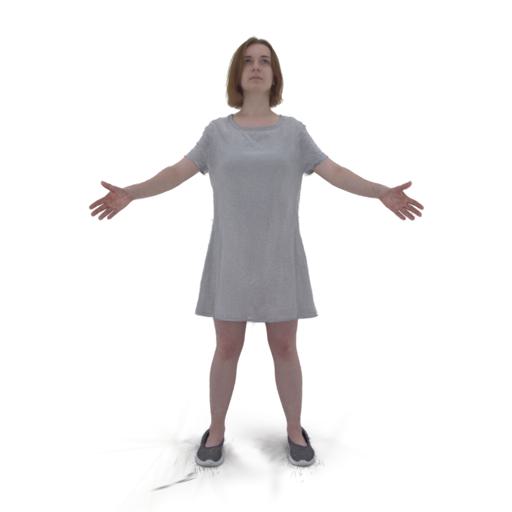}
        \end{minipage}
        \par\nointerlineskip
        \begin{minipage}[c]{\linewidth}
            \centering
            \includegraphics[width=0.8\linewidth]{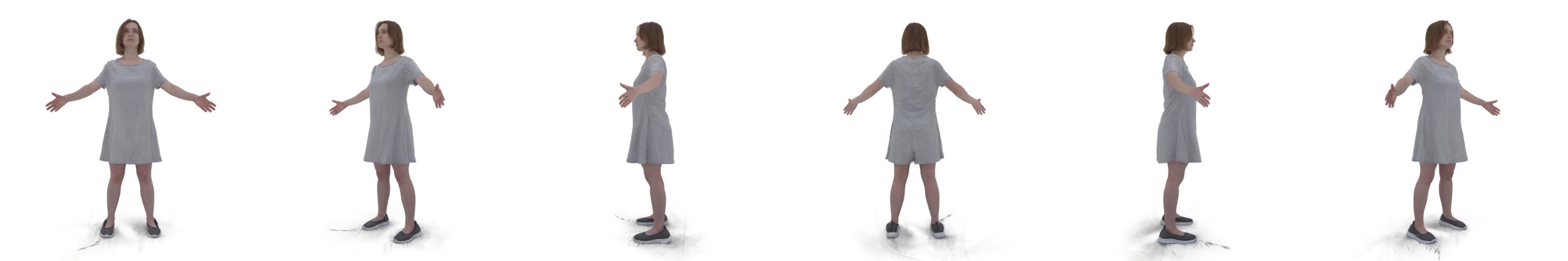}
        \end{minipage}
        \begin{minipage}[c]{\linewidth}
            \centering
            \includegraphics[width=0.8\linewidth]{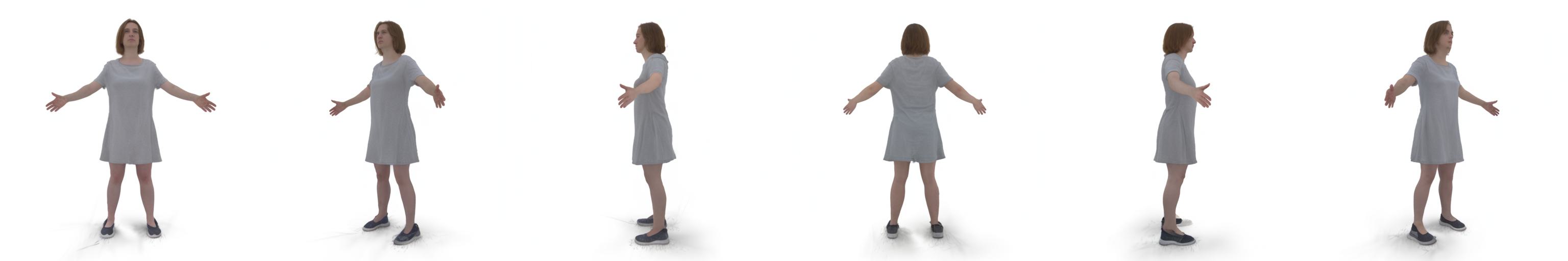}
        \end{minipage}
        \par\nointerlineskip
        \begin{minipage}[c]{\linewidth}
            \centering
            \includegraphics[width=0.8\linewidth]{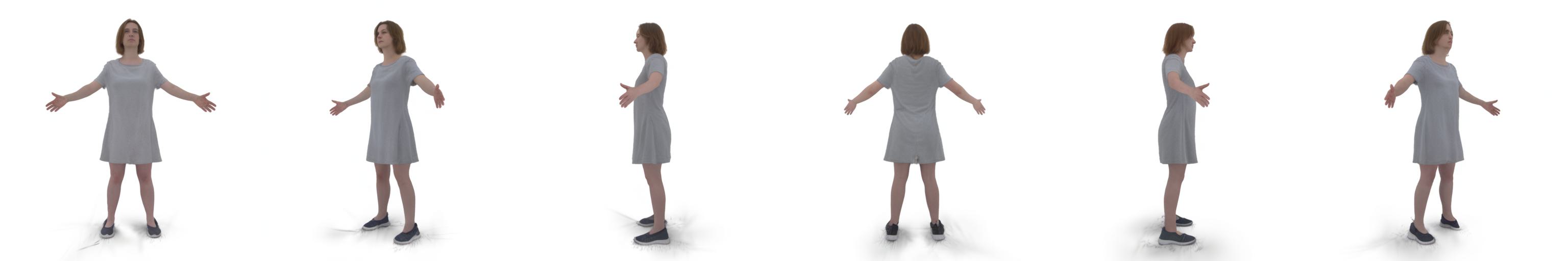}
        \end{minipage}

    \end{minipage}
    
    \caption{Qualitative results of our method on the PSGS~\cite{psgs} dataset for human body reconstruction. We show the canonical views produced by our method, compared to the GT and the input views.}
    \label{fig:body_nvs}
\end{figure*}

\begin{figure*}[ht]
    \centering
        
                \begin{minipage}[c]{0.22\linewidth}
        \centering
        \raisebox{0.25cm}{\smash{\small GT}}
    \end{minipage}%
    \hspace{0.01\linewidth}
    \begin{minipage}[b]{0.76\linewidth}
        \centering
        \includegraphics[width=1\linewidth]{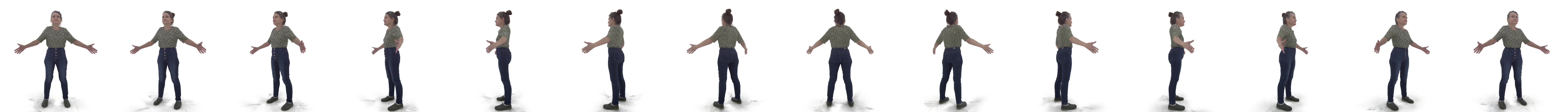}
    \end{minipage}
    \begin{minipage}[c]{0.22\linewidth}
        \centering
        \raisebox{0.25cm}{\smash{\footnotesize \textit{Ours}}}
    \end{minipage}%
    \hspace{0.01\linewidth}
    \begin{minipage}[b]{0.76\linewidth}
        \centering
        \includegraphics[width=1\linewidth]{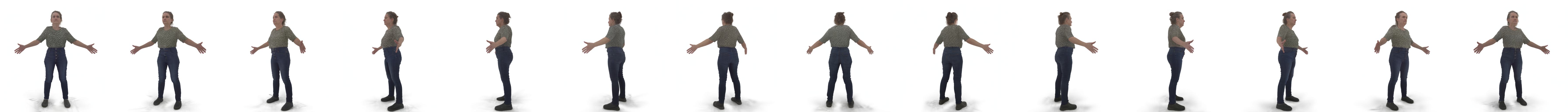}
    \end{minipage}
    \begin{minipage}[c]{0.22\linewidth}
        \centering
        \raisebox{0.25cm}{\smash{\footnotesize \textit{Ours +~\cite{psgs}}}}
    \end{minipage}%
    \hspace{0.01\linewidth}
    \begin{minipage}[b]{0.76\linewidth}
        \centering
        \includegraphics[width=1\linewidth]{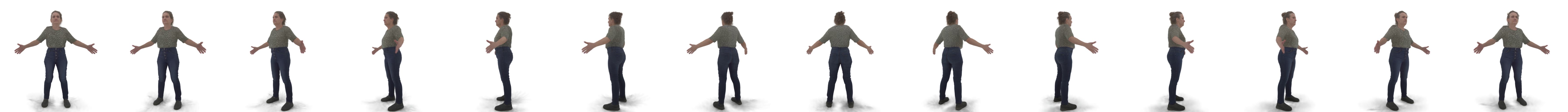}
    \end{minipage}
    \begin{minipage}[c]{0.22\linewidth}
        \centering
        \raisebox{0.25cm}{\smash{\footnotesize \textit{Ours MO +~\cite{psgs}}}}
    \end{minipage}%
    \hspace{0.01\linewidth}
    \begin{minipage}[b]{0.76\linewidth}
        \centering
        \includegraphics[width=1\linewidth]{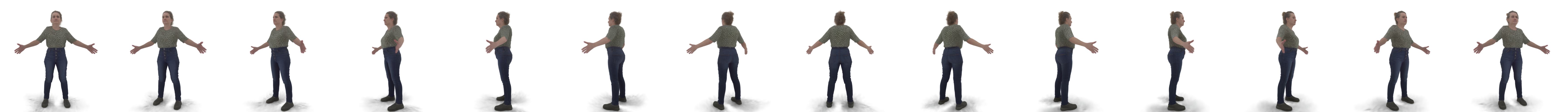}
    \end{minipage}
    \par\vspace{2mm}

    \begin{minipage}[c]{0.22\linewidth}
        \centering
        \raisebox{0.25cm}{\smash{\small GT}}
    \end{minipage}%
    \hspace{0.01\linewidth}
    \begin{minipage}[b]{0.76\linewidth}
        \centering
        \includegraphics[width=1\linewidth]{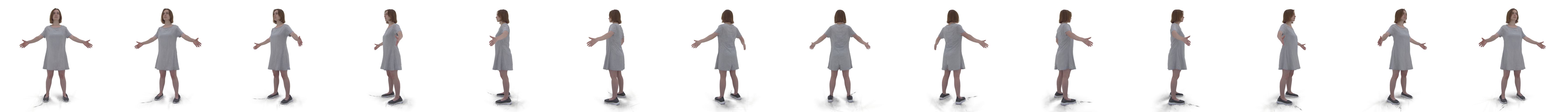}
    \end{minipage}
    \begin{minipage}[c]{0.22\linewidth}
        \centering
        \raisebox{0.25cm}{\smash{\footnotesize \textit{Ours}}}
    \end{minipage}%
    \hspace{0.01\linewidth}
    \begin{minipage}[b]{0.76\linewidth}
        \centering
        \includegraphics[width=1\linewidth]{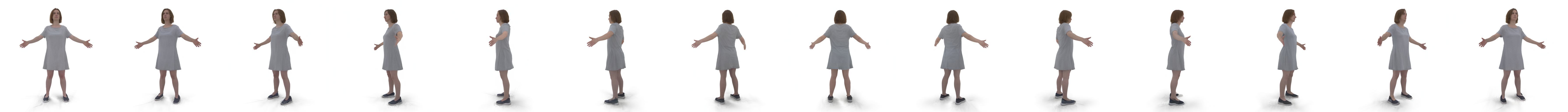}
    \end{minipage}
    \begin{minipage}[c]{0.22\linewidth}
        \centering
        \raisebox{0.25cm}{\smash{\footnotesize \textit{Ours +~\cite{psgs}}}}
    \end{minipage}%
    \hspace{0.01\linewidth}
    \begin{minipage}[b]{0.76\linewidth}
        \centering
        \includegraphics[width=1\linewidth]{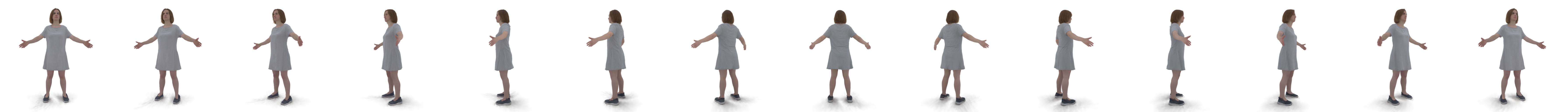}
    \end{minipage}
    \begin{minipage}[c]{0.22\linewidth}
        \centering
        \raisebox{0.25cm}{\smash{\footnotesize \textit{Ours MO +~\cite{psgs}}}}
    \end{minipage}%
    \hspace{0.01\linewidth}
    \begin{minipage}[b]{0.76\linewidth}
        \centering
        \includegraphics[width=1\linewidth]{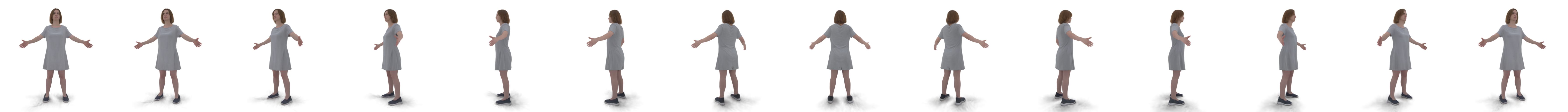}
    \end{minipage}

    \caption{Qualitative results of our method on the PSGS~\cite{psgs} dataset for human body reconstruction. We show the final views rendered after the GS-LRM 3D lifting process, compared to the GT. Since \textit{Ours} is not trained to produce only canonical views, it can also be used to render turntable views directly, without the lifting step, as shown in the second rows.}
    \label{fig:body_turntable}
\end{figure*}

\section{Object NVS training}
While we use the weights from ArchonView~\cite{archonview} for the initialization of our models, we adapt them to the higher 512$\times$512 resolution during the first stage of our training process. We show in \cref{fig:objects} example views of the SS3D dataset we employed.

\begin{figure*}[h]
  \begin{subfigure}{0.22\textwidth}
    \includegraphics[width=\textwidth]{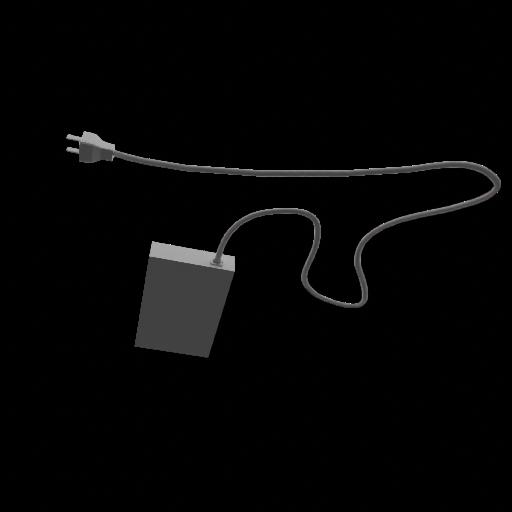}
  \end{subfigure}\hfill
  \begin{subfigure}{0.22\textwidth}
    \includegraphics[width=\textwidth]{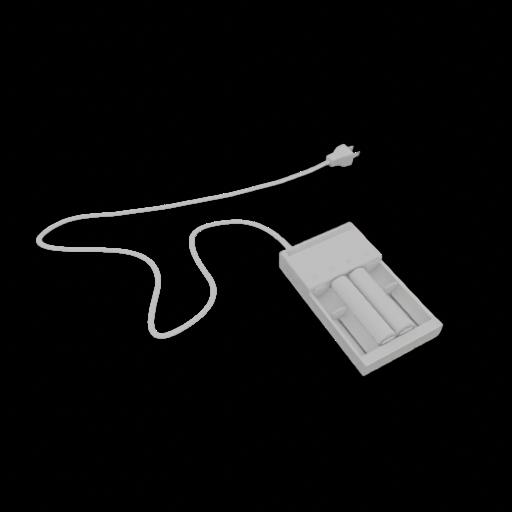}
  \end{subfigure}\hfill
  \begin{subfigure}{0.22\textwidth}
    \includegraphics[width=\textwidth]{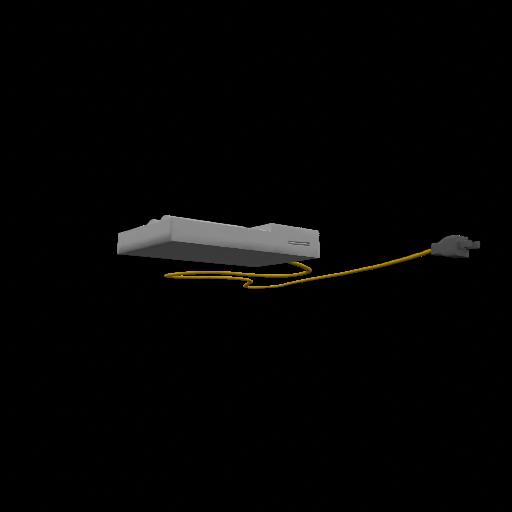}
  \end{subfigure}\hfill
  \begin{subfigure}{0.22\textwidth}
    \includegraphics[width=\textwidth]{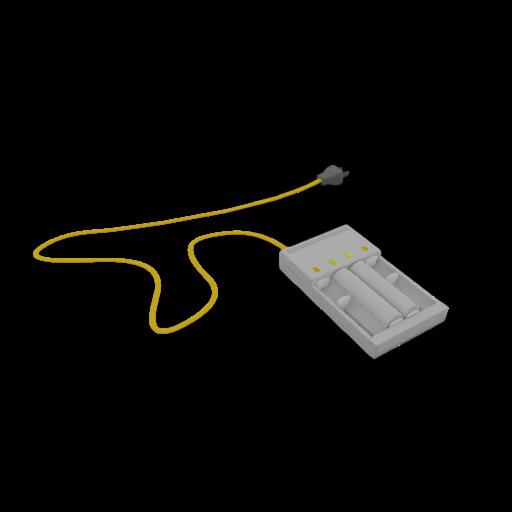}
  \end{subfigure}
  
  \caption{Example object views from the SS3D dataset.}
  \label{fig:objects}
\end{figure*}

\section{Multi-scale VQ-VAE fine-tuning}
While fine-tuning an existing component is not a core contribution of this work, we find that fine-tuning the VQ-VAE on the new resolution and dataset can further improve the results of our method, with very little cost. Complementing the ablation study in the main paper, we show in \cref{fig:vqvae-finetuning} the impact of fine-tuning the VQ-VAE on PSGS~\cite{psgs}. The results are obtained by simply autoencoding the 6 canonical views of the validation set, and show small but clear improvements in the quality and identity preservation of the reconstructed images.
\newsavebox{\vqvaefinetuningimg}
\newlength{\vqvaefinetuningcolwidth}
  

\begin{figure}[ht]
    \centering

    \begin{minipage}[c]{0.18\linewidth}
        \centering
        \footnotesize Base 256
    \end{minipage}%
    \hspace{0.01\linewidth}
    \begin{minipage}[c]{0.8\linewidth}
        \centering
        \includegraphics[width=1\linewidth]{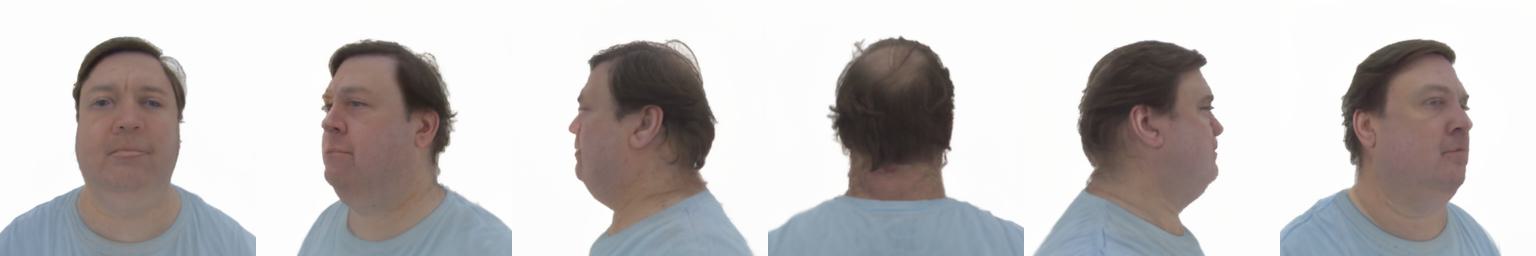}
    \end{minipage}
    \par\vspace{1mm}

    \begin{minipage}[c]{0.18\linewidth}
        \centering
        \footnotesize Fine-tuned 256
    \end{minipage}%
    \hspace{0.01\linewidth}
    \begin{minipage}[c]{0.8\linewidth}
        \centering
        \includegraphics[width=1\linewidth]{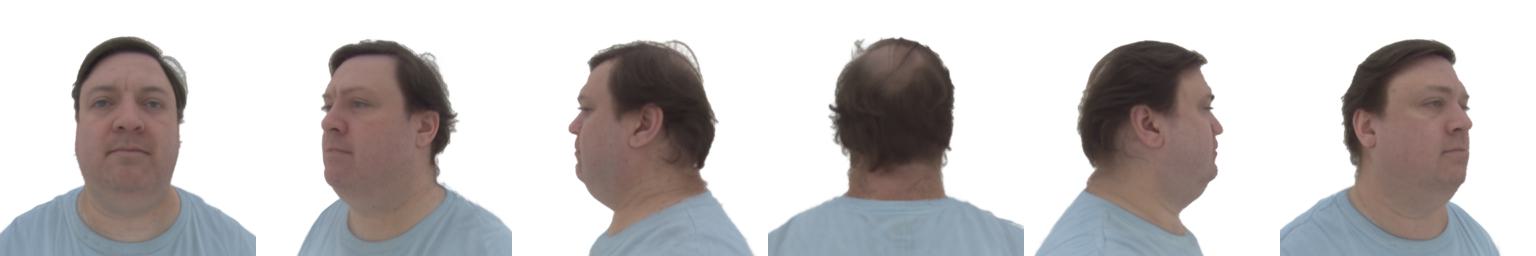}
    \end{minipage}
    \par\vspace{1mm}

    \begin{minipage}[c]{0.18\linewidth}
        \centering
        \footnotesize Base 512
    \end{minipage}%
    \hspace{0.01\linewidth}
    \begin{minipage}[c]{0.8\linewidth}
        \centering
        \includegraphics[width=1\linewidth]{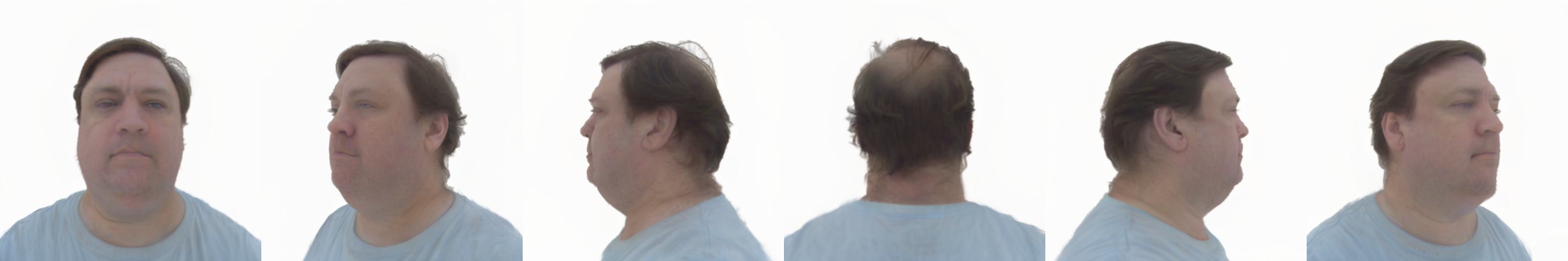}
    \end{minipage}
    \par\vspace{1mm}

    \begin{minipage}[c]{0.18\linewidth}
        \centering
        \footnotesize Fine-tuned 512
    \end{minipage}%
    \hspace{0.01\linewidth}
    \begin{minipage}[c]{0.8\linewidth}
        \centering
        \includegraphics[width=1\linewidth]{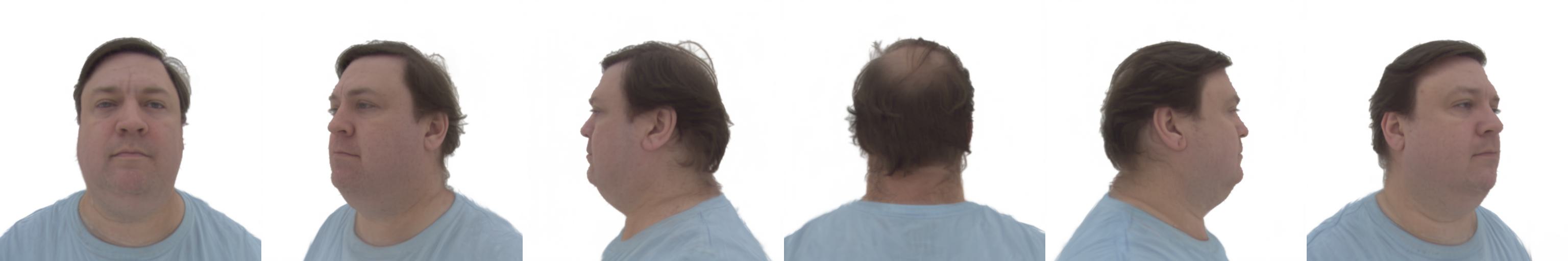}
    \end{minipage}
    \par\vspace{1mm}

    \begin{minipage}[c]{0.18\linewidth}
        \centering
        \footnotesize GT
    \end{minipage}%
    \hspace{0.01\linewidth}
    \begin{minipage}[c]{0.8\linewidth}
        \centering
        \includegraphics[width=1\linewidth]{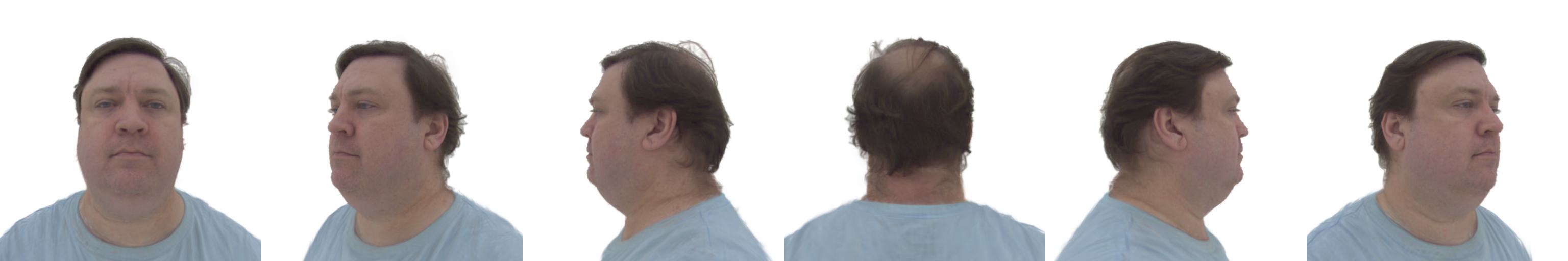}
    \end{minipage}

    \caption{Multi-scale VQ-VAE finetuning results. We show the reconstructions from the base and finetuned VQ-VAEs at both 256 and 512 resolutions, compared to the GT, showing small but noticeable improvements in reconstruction quality after finetuning.}
  \label{fig:vqvae-finetuning}
\end{figure}

\section{Additional qualitatives}
\begin{figure}[ht]
    \centering
        \begin{minipage}[c]{0.18\linewidth}
            \centering
            \small Inputs
        \end{minipage}%
        \hspace{0.01\linewidth}
        \begin{minipage}[c]{0.8\linewidth}
            \includegraphics[width=0.28\linewidth]{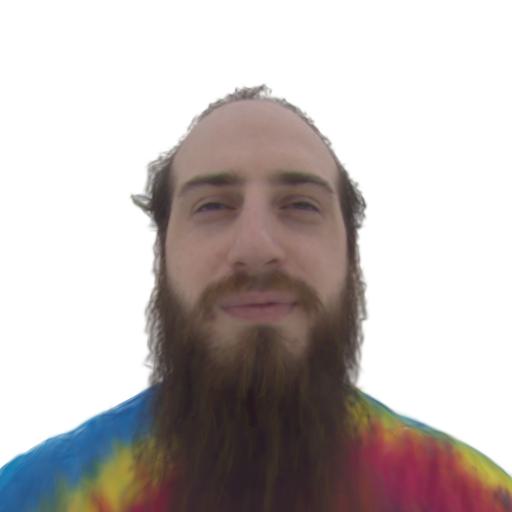}
            \hspace{0.0125\linewidth}
            \vrule width 1pt height 0.2\linewidth
            \hspace{0.0125\linewidth}
            \includegraphics[width=0.28\linewidth]{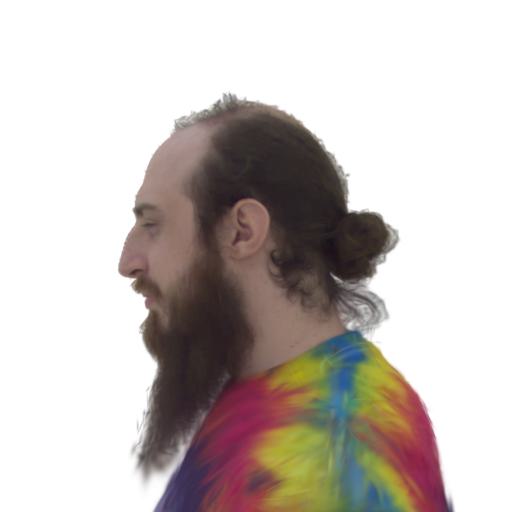}
            \hspace{0.0125\linewidth}
            \includegraphics[width=0.28\linewidth]{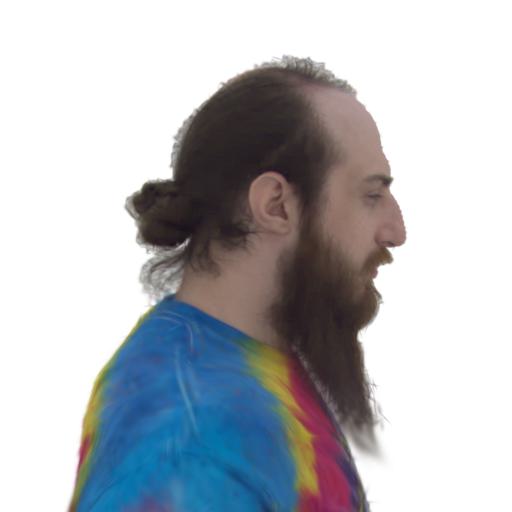}
        \end{minipage}

        \par\nointerlineskip

                \begin{minipage}[c]{0.18\linewidth}
        \centering
        \scriptsize GT
    \end{minipage}%
    \hspace{0.01\linewidth}
    \begin{minipage}[c]{0.8\linewidth}
        \centering
        \includegraphics[width=1\linewidth]{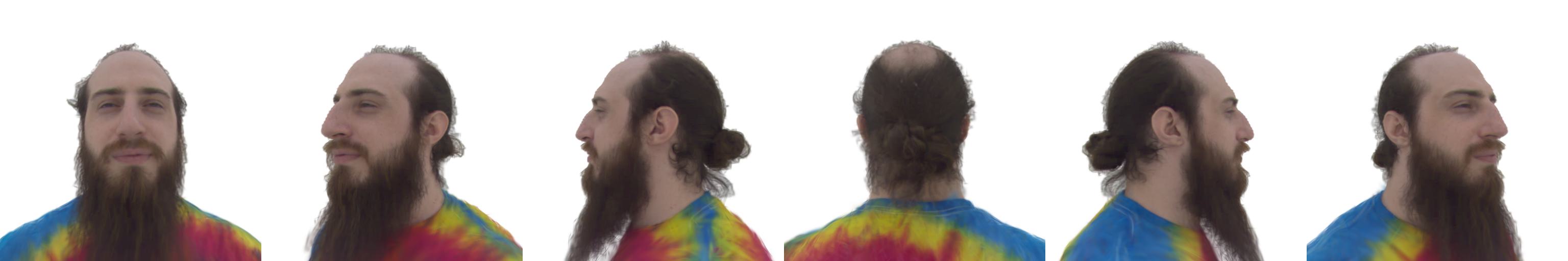}
    \end{minipage}
    \par\nointerlineskip
    \begin{minipage}[c]{0.18\linewidth}
        \centering
        \scriptsize SI (PSGS) I~\cite{splatterimage}
    \end{minipage}%
    \hspace{0.01\linewidth}
    \begin{minipage}[c]{0.8\linewidth}
        \centering
        \includegraphics[width=1\linewidth]{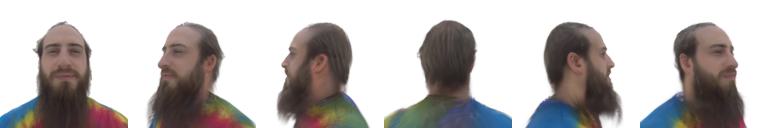}
    \end{minipage}
    \par\nointerlineskip
    \begin{minipage}[c]{0.18\linewidth}
        \centering
        \scriptsize SI (PSGS) II~\cite{splatterimage}
    \end{minipage}%
    \hspace{0.01\linewidth}
    \begin{minipage}[c]{0.8\linewidth}
        \centering
        \includegraphics[width=1\linewidth]{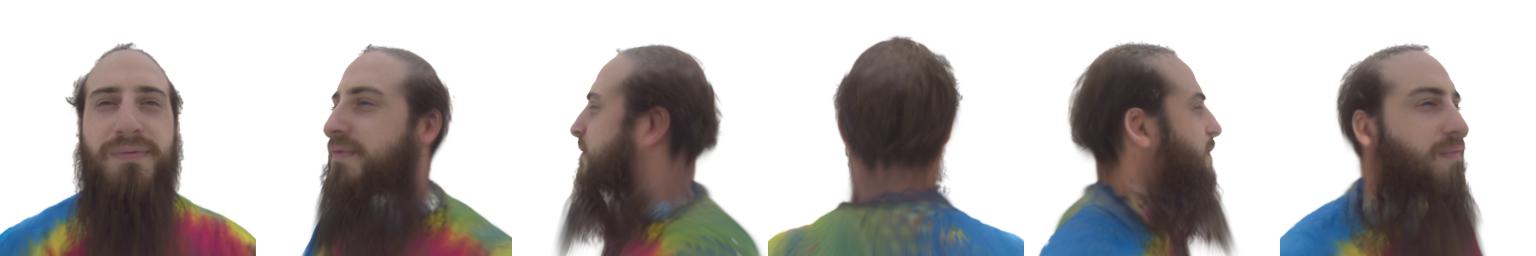}
    \end{minipage}
    \par\nointerlineskip
    \begin{minipage}[c]{0.18\linewidth}
        \centering
        \scriptsize FaceLift~\cite{facelift}
    \end{minipage}%
    \hspace{0.01\linewidth}
    \begin{minipage}[c]{0.8\linewidth}
        \centering
        \includegraphics[width=1\linewidth]{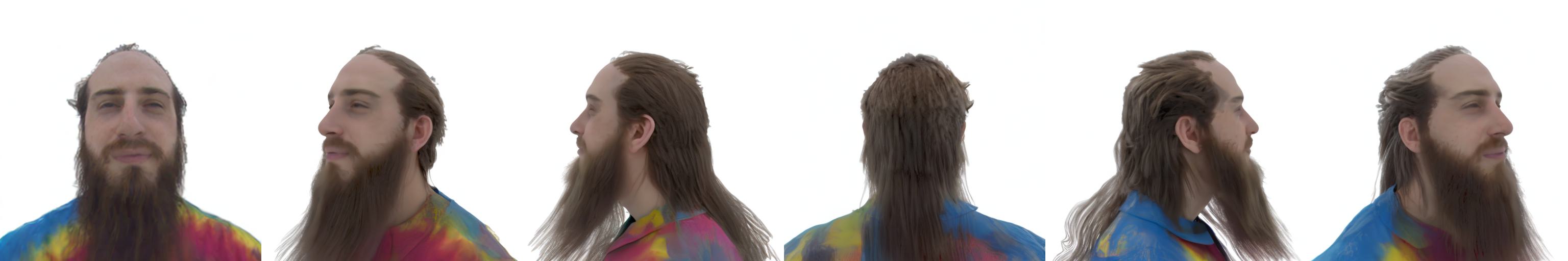}
    \end{minipage}
    \par\nointerlineskip
    \begin{minipage}[c]{0.18\linewidth}
        \centering
        \scriptsize FaceLift (PSGS)~\cite{facelift}
    \end{minipage}%
    \hspace{0.01\linewidth}
    \begin{minipage}[c]{0.8\linewidth}
        \centering
        \includegraphics[width=1\linewidth]{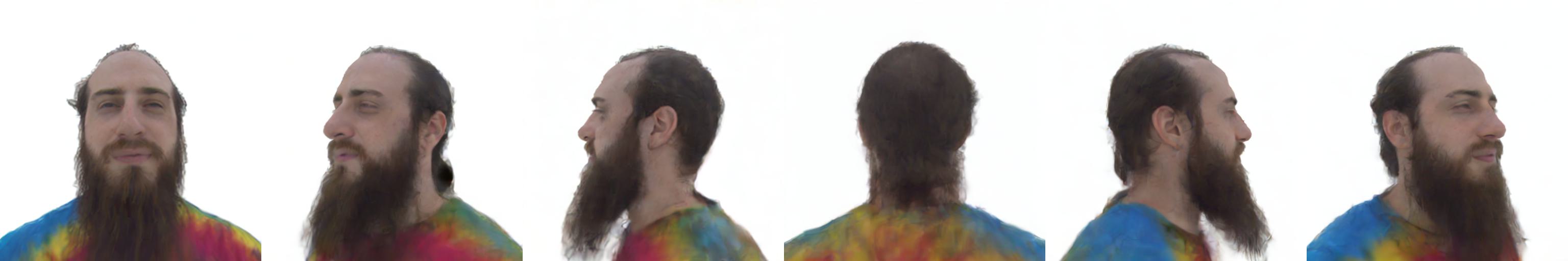}
    \end{minipage}
    \par\nointerlineskip
    \begin{minipage}[c]{0.18\linewidth}
        \centering
        \scriptsize \textit{Ours}
    \end{minipage}%
    \hspace{0.01\linewidth}
    \begin{minipage}[c]{0.8\linewidth}
        \centering
        \includegraphics[width=1\linewidth]{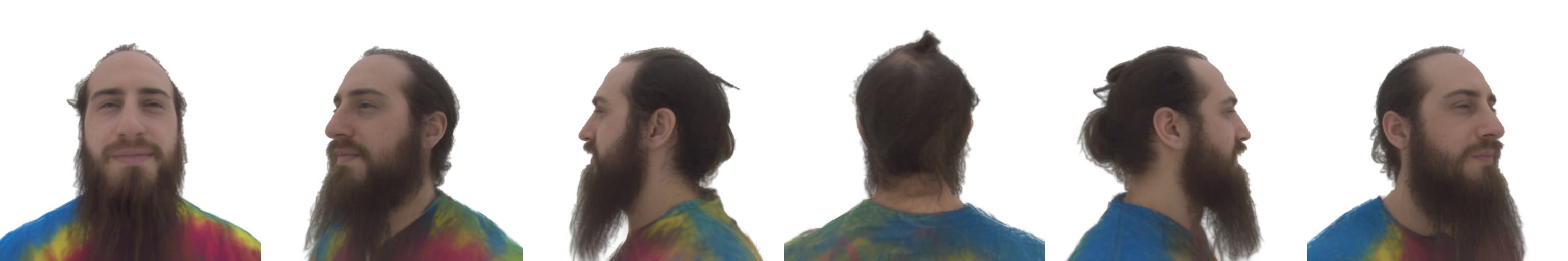}
    \end{minipage}
    \par\nointerlineskip
    \begin{minipage}[c]{0.18\linewidth}
        \centering
        \scriptsize \textit{Ours MO}
    \end{minipage}%
    \hspace{0.01\linewidth}
    \begin{minipage}[c]{0.8\linewidth}
        \centering
        \includegraphics[width=1\linewidth]{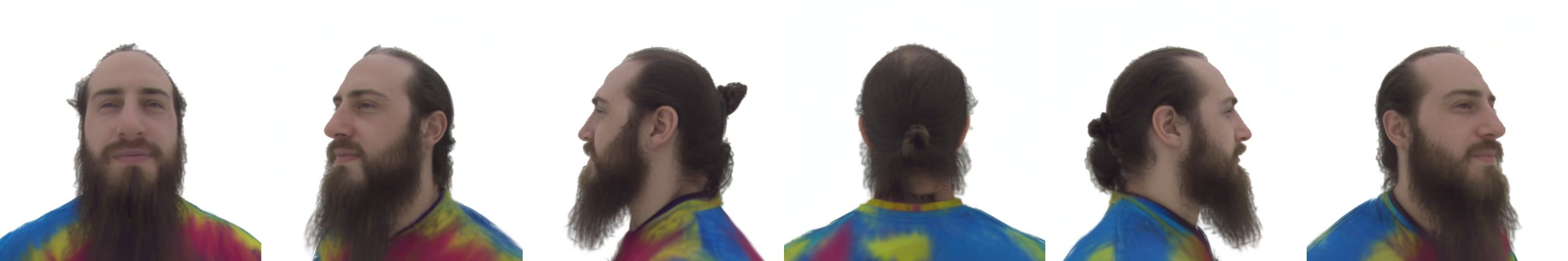}
    \end{minipage}
    \par\nointerlineskip
    \begin{minipage}[c]{0.18\linewidth}
        \centering
        \scriptsize \textit{Ours MI-MO}
    \end{minipage}%
    \hspace{0.01\linewidth}
    \begin{minipage}[c]{0.8\linewidth}
        \centering
        \includegraphics[width=1\linewidth]{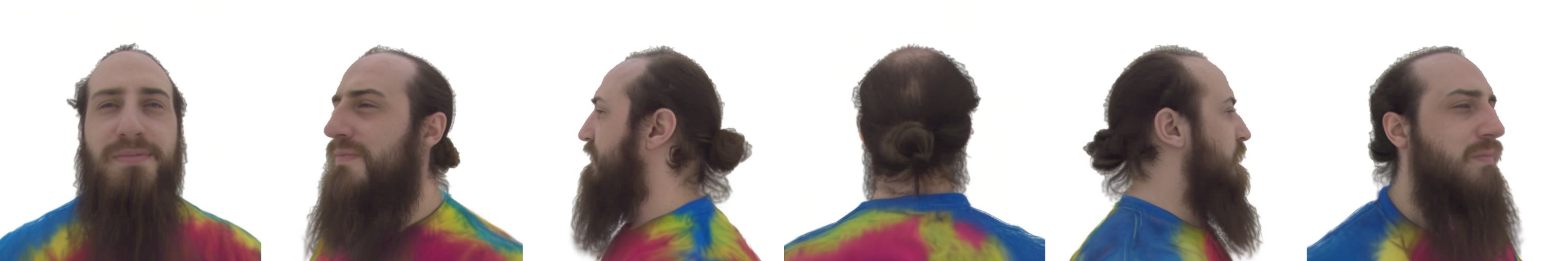}
    \end{minipage}
    \caption{Qualitative results on PSGS~\cite{psgs} validation set on the 6 canonical views for the NVS part of each model.}
    \label{fig:turntable1-nvs}
\end{figure}

\begin{figure*}[ht]
    \centering
        \begin{minipage}[c]{0.22\linewidth}
            \centering
            \small Inputs
        \end{minipage}%
        \hspace{0.01\linewidth}
        \begin{minipage}[c]{0.76\linewidth}
            \includegraphics[width=0.13\linewidth]{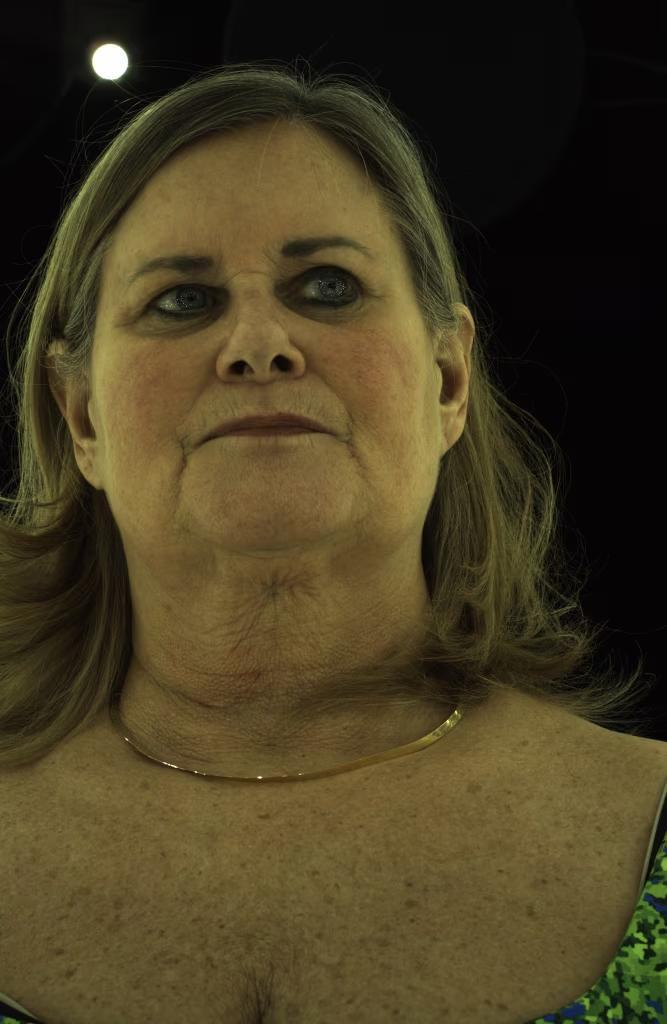}
            \hspace{0.01\linewidth}
            \vrule width 1pt height 0.2\linewidth
            \hspace{0.01\linewidth}
            \includegraphics[width=0.13\linewidth]{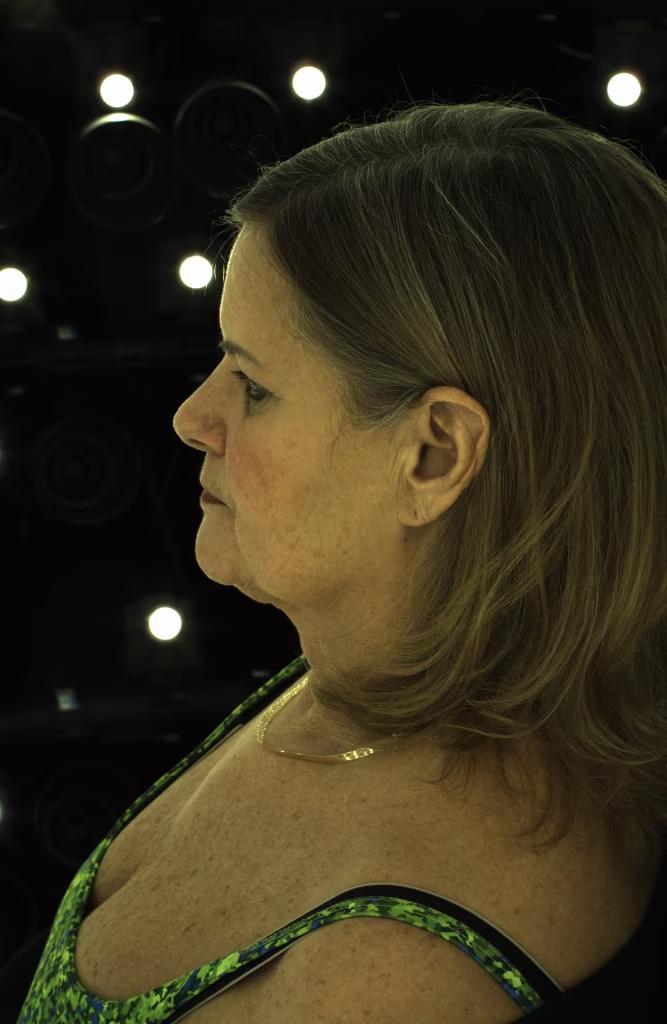}
            \hspace{0.01\linewidth}
            \includegraphics[width=0.13\linewidth]{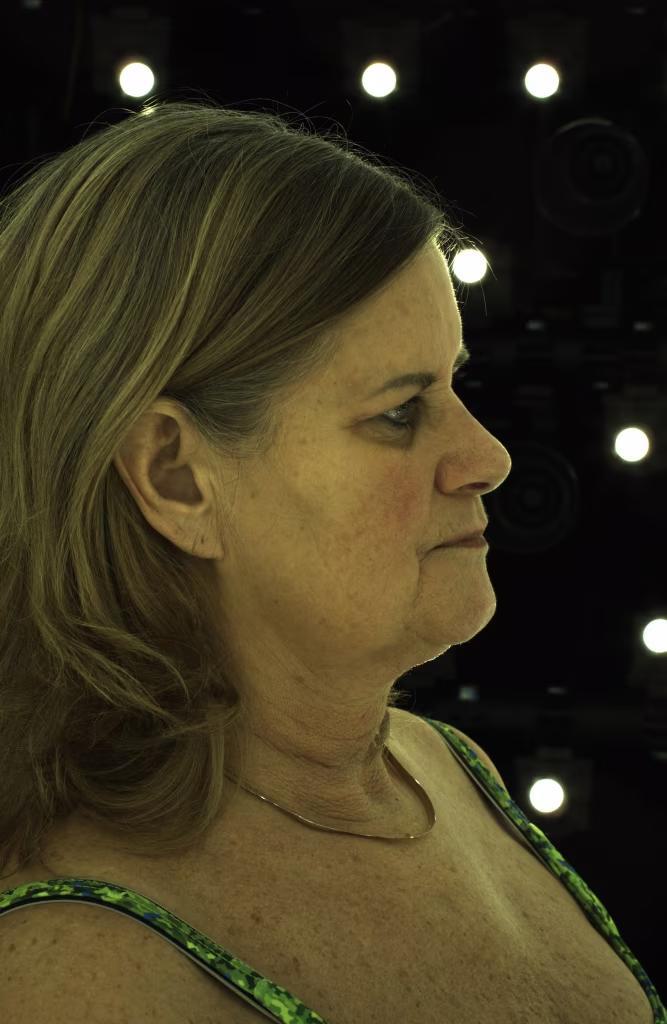}
        \end{minipage}

        \par\vspace{1mm}

                \begin{minipage}[c]{0.22\linewidth}
        \centering
        \scriptsize GT
    \end{minipage}%
    \hspace{0.01\linewidth}
    \begin{minipage}[c]{0.76\linewidth}
        \centering
        \includegraphics[width=1\linewidth]{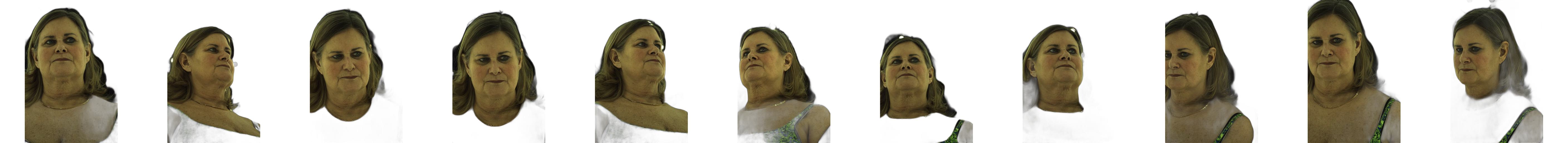}
    \end{minipage}
    \par\vspace{1mm}
    \begin{minipage}[c]{0.22\linewidth}
        \centering
        \scriptsize SI (PSGS)~\cite{splatterimage}
    \end{minipage}%
    \hspace{0.01\linewidth}
    \begin{minipage}[c]{0.76\linewidth}
        \centering
        \includegraphics[width=1\linewidth]{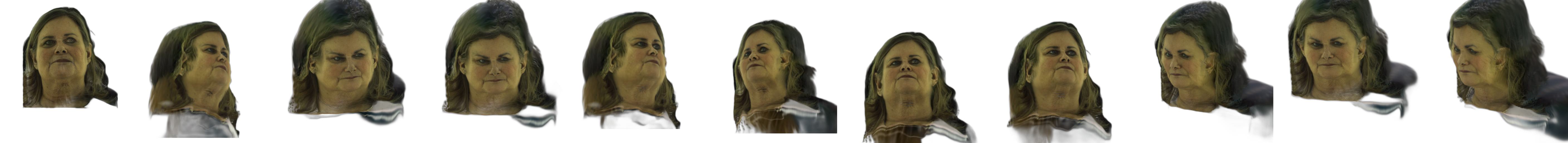}
    \end{minipage}
    \par\vspace{1mm}
    \begin{minipage}[c]{0.22\linewidth}
        \centering
        \scriptsize FaceLift~\cite{facelift}
    \end{minipage}%
    \hspace{0.01\linewidth}
    \begin{minipage}[c]{0.76\linewidth}
        \centering
        \includegraphics[width=1\linewidth]{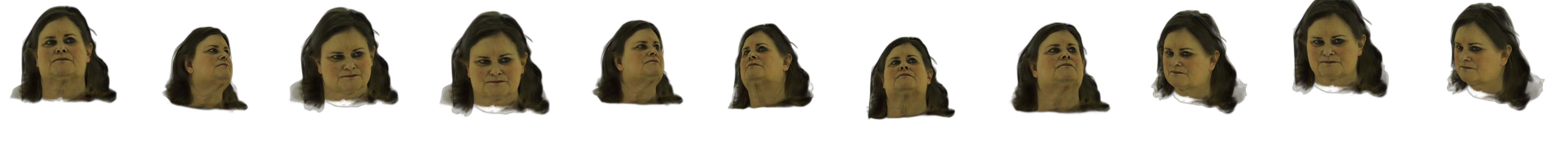}
    \end{minipage}
    \par\vspace{1mm}
    \begin{minipage}[c]{0.22\linewidth}
        \centering
        \scriptsize FaceLift (PSGS)~\cite{facelift}
    \end{minipage}%
    \hspace{0.01\linewidth}
    \begin{minipage}[c]{0.76\linewidth}
        \centering
        \includegraphics[width=1\linewidth]{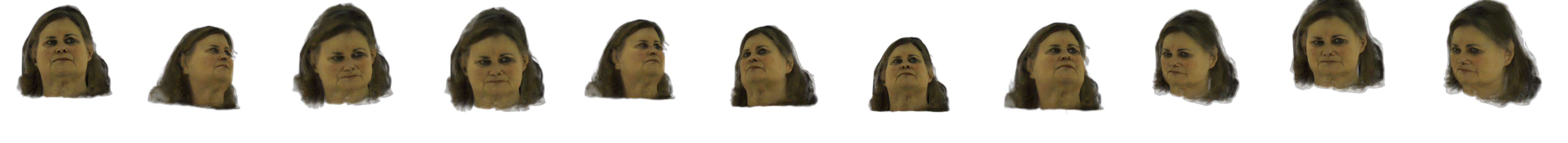}
    \end{minipage}
    \par\vspace{1mm}
    \begin{minipage}[c]{0.22\linewidth}
        \centering
        \scriptsize \textit{Ours +~\cite{psgs}}
    \end{minipage}%
    \hspace{0.01\linewidth}
    \begin{minipage}[c]{0.76\linewidth}
        \centering
        \includegraphics[width=1\linewidth]{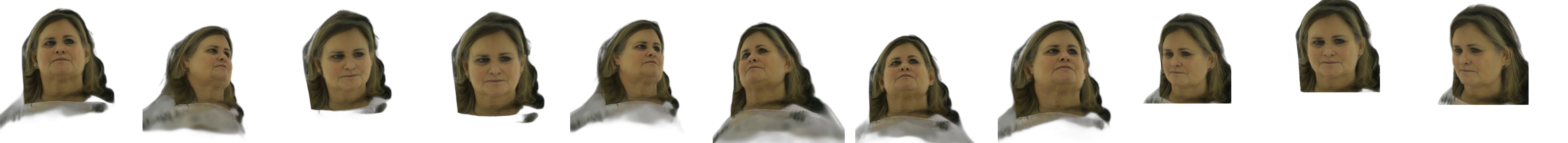}
    \end{minipage}
    \par\vspace{1mm}
    \begin{minipage}[c]{0.22\linewidth}
        \centering
        \scriptsize \textit{Ours MO +~\cite{psgs}}
    \end{minipage}%
    \hspace{0.01\linewidth}
    \begin{minipage}[c]{0.76\linewidth}
        \centering
        \includegraphics[width=1\linewidth]{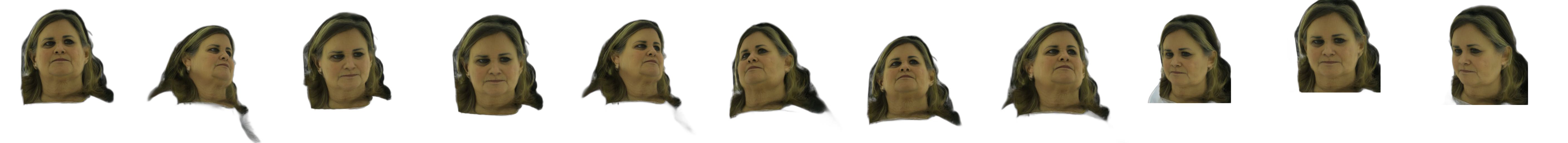}
    \end{minipage}
    \par\vspace{1mm}
    \begin{minipage}[c]{0.22\linewidth}
        \centering
        \scriptsize \textit{Ours MI-MO +~\cite{psgs}}
    \end{minipage}%
    \hspace{0.01\linewidth}
    \begin{minipage}[c]{0.76\linewidth}
        \centering
        \includegraphics[width=1\linewidth]{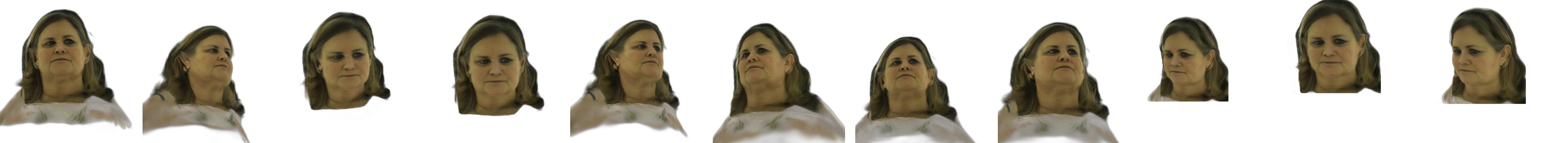}
    \end{minipage}
    \caption{Qualitative results on AVA-256~\cite{ava256} for subject 20220310--1128--ZSC414}
    \label{fig:ava2}
\end{figure*}
We show in \cref{fig:turntable1-nvs} the canonical views produced by our method for the subject shown in \cref{fig:turntable1} of the main paper. We also show additional qualitative results on the AVA-256~\cite{ava256} dataset in \cref{fig:ava2}.
  \clearpage
\else
  \begin{abstract}
    Photorealistic novel view synthesis of people remains challenging at high spatial resolutions and across multiple target cameras, where preserving identity, fine appearance details, and geometric coherence is critical. We build on the next-scale autoregressive paradigm and adapt it for human-centric view synthesis by enabling higher image resolutions, multi-view outputs and stronger cross-view consistency in a single forward pass. We train on a synthetic dataset of human faces spanning diverse identities and apparel. Contrary to diffusion models, this paradigm does not need 2D pre-training and, thanks to its next-scale architecture, it benefits from lower-resolution, general-purpose pre-trainings, with the full-sized purpose-specific images being used only in the last training stages. This enables our architecture to converge with a smaller amount of purpose-specific training data, allowing us to use a smaller but more realistic training dataset. The resulting model produces sharp and realistic views, with the option to synthesize multiple novel viewpoints simultaneously for improved agreement across views.
    Empirically, we observe gains in perceptual fidelity and cross-view coherence on human subjects, demonstrating that next-scale autoregression is an effective backbone for scalable, multi-output human view synthesis. We also couple our pipeline with an existing transformer-based model for pixel-aligned 3D gaussian lifting from multi-view facial inputs, resulting in accurate and photorealistic 3D models of human faces.
\end{abstract}

\section{Introduction}
Synthesizing photorealistic novel views of human faces and bodies remains challenging at high resolution and across wide viewpoint changes, where methods must preserve identity while maintaining geometric, shading, and appearance consistency. Recent approaches tackle complementary subproblems but exhibit trade-offs: near-frontal pipelines (e.g., Splatter Image~\cite{splatterimage} and VoluMe~\cite{volume}) can run in real time and emphasize quality around canonical views but struggle as viewpoint departs from frontal, while large-scale, heavily pre-trained image-to-3D systems (e.g., FaceLift~\cite{facelift}) achieve strong detail yet often produce artificial-looking renderings and rely on extensive synthetic data with varied lighting and expressions. 
We pursue a different path: extending a next-scale autoregressive transformers (VAR~\cite{VAR}) for human-centric view synthesis that operates under data scarcity and does not necessitate of a large 2D image-generation pretraining or extensive synthetic data. Our system builds on the novel-view synthesis architecture introduced in ArchonView~\cite{archonview} and extends it to support higher resolutions ($512\times512$) and multiple simultaneous input and output views, converging with minimal purpose-specific realistic training data and achieving higher accuracy and detail in a single forward pass. While transformer-based diffusion models can naturally provide multi-view attention over multiple passes, the same is not true for autoregressive models, which operate on a single pass. Our next-scale autoregressive architecture, instead, allows tokens of different scales to attend to previous scales of every output view, achieving cross-view attention in a single forward pass, with faster speeds compared to diffusion models.
This design prioritizes realism, especially on off-frontal views, while enforcing identity preservation and geometric coherence across simultaneously synthesized cameras. In head-to-head comparisons on our validation set and a subset of images curated by prior work, our approach delivers more realistic results than competing methods, which can fail to converge correctly when training with naturally scarce realistic data, highlighting the importance of our architectural choices and training scheme.
As pre-trained next-scale autoregressive backbones such as~\cite{Infinity,InfinityStar} become more available, akin to diffusion models, our approach offers a practical foundation for controllable, consistent multi-camera rendering of people that can capitalize on stronger initializations without sacrificing its core strength: convergence and realism in low-data regimes.

\begin{figure}[t]
    \centering
    \begin{tabular*}{\textwidth}{@{\extracolsep{\fill}} ccccccc }
        \includegraphics[width=0.15\textwidth]{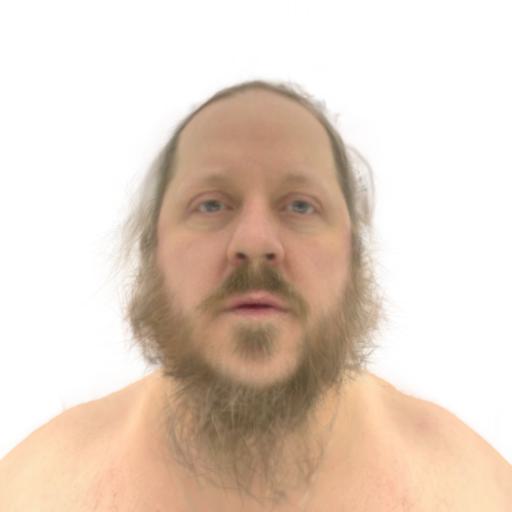} &
        \includegraphics[width=0.15\textwidth]{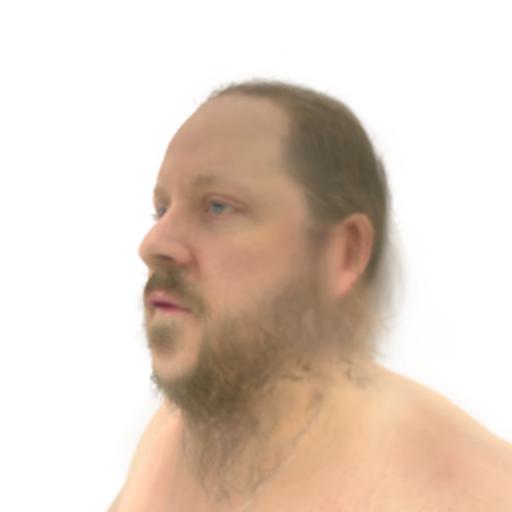} &
        \includegraphics[width=0.15\textwidth]{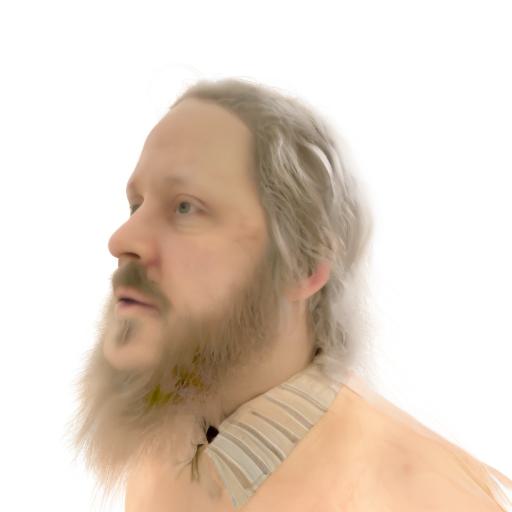} &
        \includegraphics[width=0.15\textwidth]{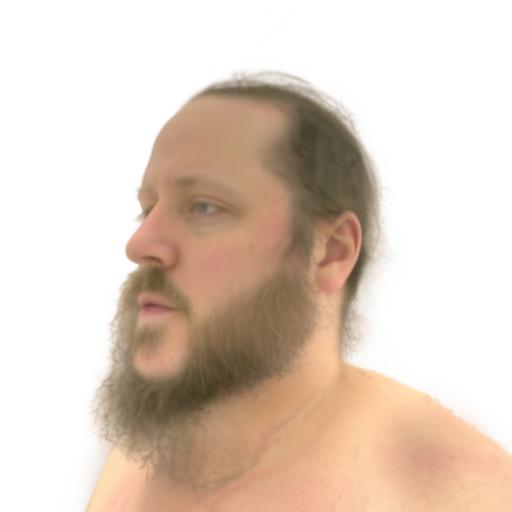} &
        \includegraphics[width=0.15\textwidth]{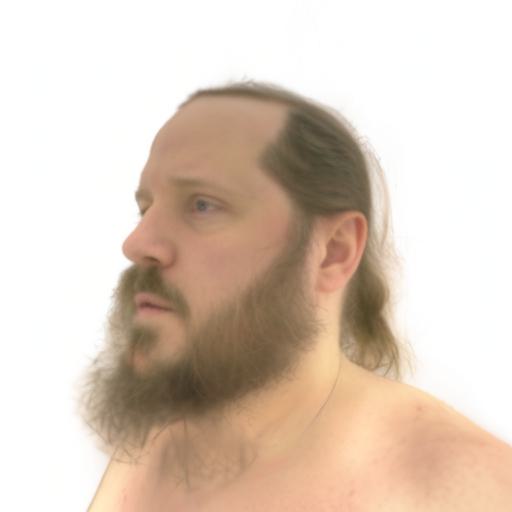} &
        \includegraphics[width=0.15\textwidth]{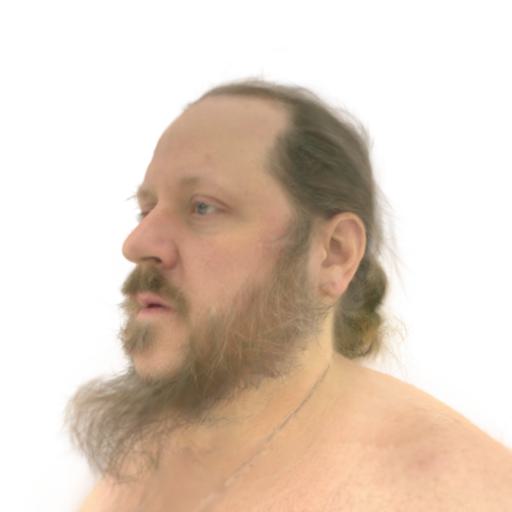} \\
        Input & Splatter~\cite{splatterimage} & FaceLift~\cite{facelift} & \textit{Ours MO} & \textit{Ours MI-MO$^\dagger$} & GT \\
    \end{tabular*}
    \caption{Face reconstruction from single view: our method achieves more realistic results than competing methods. $^\dagger$Variant of our method trained to handle multiple loosely-posed inputs, it achieves even better results thanks to the additional inputs.}
    \label{fig:teaser}
\end{figure}

To summarize our contributions:
\begin{itemize}
    \item We extend the next-scale autoregressive transformer architecture for multi-view human face novel view synthesis, supporting multiple simultaneous inputs and outputs;
    \item We introduce a training strategy that enables the model to converge at high resolution (512$\times$512) without the need to re-train the model from scratch, thus lowering the amount of needed compute and data;
    \item We couple our pipeline with an existing transformer-based model for pixel-aligned 3D gaussian lifting~\cite{psgs}, and we benchmark the results against state-of-the-art models.
\end{itemize}

Finally, we showcase the flexibility of our method by applying it to human body reconstruction without any architectural changes in the supplementary material, showing a potential direction for future work.

  \section{Related works}
\label{sec:related}

\paragraph{Generative models and novel view synthesis.}
Diffusion models have shown impressive results in generating high-quality images~\cite{Peebles2022DiT, ho2021classifierfree, rombach2021highresolution}. Autoregressive approaches have been used for image generation as well~\cite{pmlr-v119-chen20s, ramesh2021zeroshottexttoimagegeneration, razavi2019generating, yu2022scaling,VQGAN}, but they have been largely outperformed by diffusion models in terms of image quality and diversity. Recently, the next-scale transformer architecture has been proposed for high-quality image generation~\cite{VAR}, changing the traditional raster-scan order of autoregressive models to a scale-based order, which allows for better modeling and speed, outperforming diffusion models~\cite{VAR} and originating a new research direction. It has been developed and applied to multiple tasks~\cite{ren2025flowar, ren2024mvardecoupledscalewiseautoregressive, tang2025hart, zhang2024varclip, chen2024sar3d, qu2025visual}, including novel view synthesis of objects~\cite{archonview}, demonstrating superior performance compared to similarly-trained diffusion models such as Zero 1-to-3~\cite{zero1to3}, Zero 123-XL~\cite{objaverseXL} and EscherNet~\cite{eschernet}, without the need for large 2D image pre-training. However, the need for large 3D datasets, generally unavailable for human faces and bodies, has limited its application to these domains. We propose a training strategy that maximizes the use of available data and models, unlocking the use of this powerful architecture for human face novel view synthesis. Concurrent works such as~\cite{Infinity,InfinityStar} have further improved the next-scale transformer architecture, so they represent a promising direction for future work to further improve the results of our method. 

\paragraph{Face reconstruction.}
Face reconstruction has been a long-standing problem in computer vision, with applications in various fields such as virtual reality, gaming, and human-computer interaction. Traditional methods for face reconstruction include 3D morphable models (3DMM)~\cite{3dmm}, which use a parametric model to represent the shape and appearance of faces. More recently, deep learning-based approaches have been proposed for face reconstruction, such as GANs~\cite{10.5555/2969033.2969125, DBLP:conf/cvpr/KarrasLA19, 9156570}. These methods have shown promising results, but they often struggle with consistency in non-frontal views. Subsequent works such as PanoHead~\cite{an2023panohead}, Rodin~\cite{10204640} and RodinHD~\cite{zhang2025rodinhd} greatly improved the consistency of novel views, but still struggle with realism. Methods such as~\cite{10.1145/3528223.3530143,avat3r,saito2024rgca} have shown high quality results, but may require a lengthy or inaccessible capture process~\cite{volume}, sometimes also requiring posed images, which limit their applicability. FaceLift~\cite{facelift} focuses on a simpler capture scenario: it employs a two-stage approach for face reconstruction from a single frontal view, first generating 6 canonical views using a diffusion model, and then using a GS-LRM~\cite{10.1007/978-3-031-72670-5_1} to lift these views to a full 3D representation. However, the use of a diffusion model requires large-scale 2D image pre-training as well as sufficiently large dataset for face reconstruction which, given the lack of sufficiently large 3D datasets of human faces, is synthetically generated. This can limit the realism of the method. In contrast, our method converges using a relatively small dataset of 3D face scans and achieves increased realism, while also allowing the use of multiple conditioning views to further improve the consistency and quality of the generated novel views.\\
A different approach to face reconstruction is proposed by VoluMe~\cite{volume}, which is based on Splatter Image~\cite{splatterimage}, a fast 2D U-Net architecture that generates a volumetric representation of the face from a single image at real-time speed. This is particularly suitable for near-frontal views, as the method predicts the 3D representation directly from the input image, and it finds applications in video conferencing scenarios, but it struggles with detail and realism in non-frontal views. In contrast, our method is suitable for generating high-quality views from a wide range of angles, including non-frontal views.\\
Finally, a different line of works focuses on reconstructing a full body model from one or more images, such as LHM~\cite{LHM}, IDOL~\cite{IDOL} and PF-LHM~\cite{PFLHM3A}. We do not focus on full-body reconstruction in this work, but we qualitatively show that our method can be applied to this domain without any architectural changes, showing the flexibility of the method and a potential direction for future work.

\section{Method}
\label{sec:method}
Similarly to FaceLift~\cite{facelift}, we approach the problem of modeling photorealistic human head avatars by decomposing it in two stages. In the first stage, we generate 6 specific ``canonical'' views of the head from an input conditioning, at angles 0, 45, 90, 180, 270, 315 degrees. In the second stage, we use an existing model to reconstruct the full 3D head from these views. The pipeline does not need the camera poses of the input images, since the only poses needed throughout the process are the fixed canonical ones. Contrary to FaceLift, which works only with one input view, our pipeline also accepts three loosely posed input images (front and side views), which provide more complete information about the subject to be reconstructed. The first stage of our pipeline is based on a novel next-scale transformer architecture named VAR~\cite{VAR}, in its novel-view synthesis variant ArchonView~\cite{archonview}. We introduce the basics of this architecture below, and then describe how we adapt it to our task, enabling it to handle multiple input views and generate multiple output views simultaneously. For the second stage, we use an existing GS-LRM named MV2Splat~\cite{psgs}.

\subsection{Image encoding}
In VAR transformers, images are encoded using a residual~\cite{lee2022autoregressive} multi-scale~\cite{VAR} variant of the original VQ-VGAN~\cite{VQGAN} architecture. The VQ-VAE encoder encodes an image $\mathbf{I} \in \mathbb{R}^{h \times w \times 3}$ into a set of discrete latent feature maps $\{\mathbf{Z}_1, \mathbf{Z}_2, \ldots, \mathbf{Z}_K\}$ at $K$ different scales, where each $\mathbf{Z}_k \in \mathbb{R}^{h_k \times w_k \times C}$ is a 2D grid of discrete tokens obtained by quantizing continuous latent features with a learned codebook, with $C$ the embedding dimension. The scales are arranged such that $(h_1, w_1) < (h_2, w_2) < \ldots < (h_K, w_K)$, allowing the model to capture both fine-grained details and global context. The codebook has size $V$, it is learned during training, and it is shared across all scales. The VQ-VAE decoder reconstructs the image from the multi-scale latent feature maps by progressively upsampling and decoding them back to the original resolution. The size of the final scale $h_K \times w_K$ and a downsampling factor control the size of the output image. The auto-encoder is trained with a compound loss which includes a perceptual and a discriminative loss, and it is frozen after the initial training.

\begin{figure*}[]
    \centering
    \includegraphics[width=1.0\textwidth]{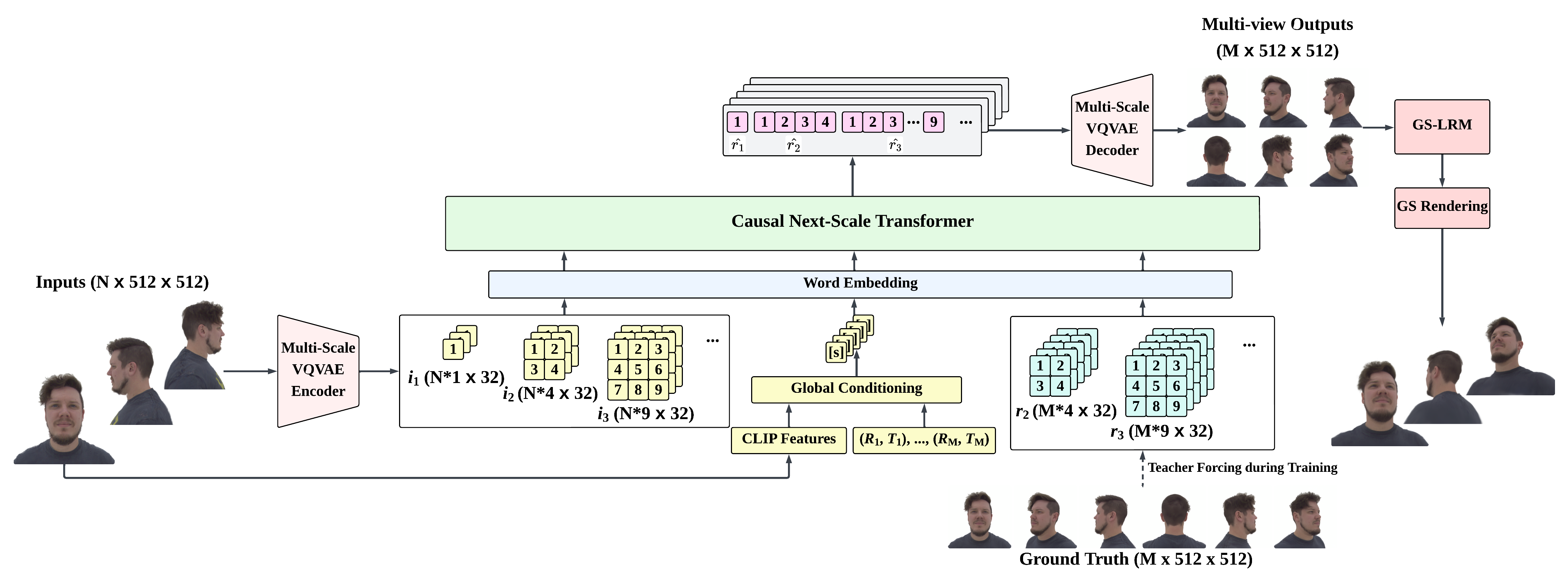}
    \vspace{-0.5cm}
    \caption{\textbf{Our architecture}: input images are encoded into multi-scale features by the VQ-VAE encoder and used as inputs for the transformer. They are also embedded into a global conditioning along with the camera-pose RT matrix, used both in the AdaLN layers and the SOS tokens. The transformer outputs one scale at a time for every output view simultaneously, up to the final scale. The output features are decoded by the VQ-VAE decoder to produce the final RGB images, which can be lifted to 3D gaussians by an off-the-shelf GS-LRM model. During training, previous scales are teacher-forced, while during inference the model is auto-regressive. The architecture is the same for all our models, with the only difference being the number of input and output views.}
    \label{fig:pipeline}
\end{figure*}
\newsavebox{\pnthirdimg}
\newlength{\pncolwidth}
\begin{figure}[h]
  \begin{subfigure}{\textwidth}
    \begin{minipage}[c]{0.25\textwidth}
      \raggedleft
      \scriptsize 256x256, 10 scales
    \end{minipage}%
    \hspace{0.5em}%
    \begin{minipage}[c]{0.75\textwidth}
      \includegraphics[height=0.68cm]{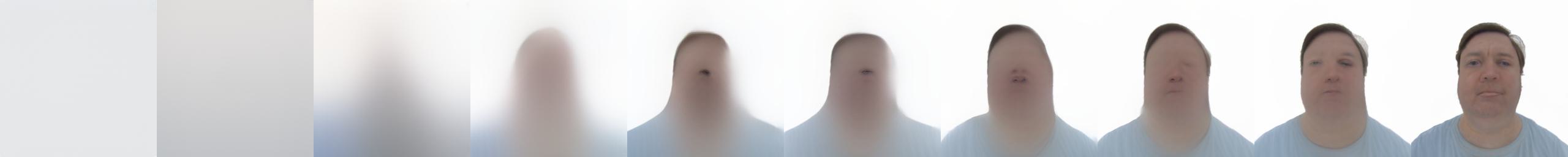}
    \end{minipage}
  \end{subfigure}
  
  \begin{subfigure}{\textwidth}
    \begin{minipage}[c]{0.25\textwidth}
      \raggedleft
      \scriptsize 512x512, 10 scales
    \end{minipage}%
    \hspace{0.5em}%
    \begin{minipage}[c]{0.75\textwidth}
      \includegraphics[height=0.68cm]{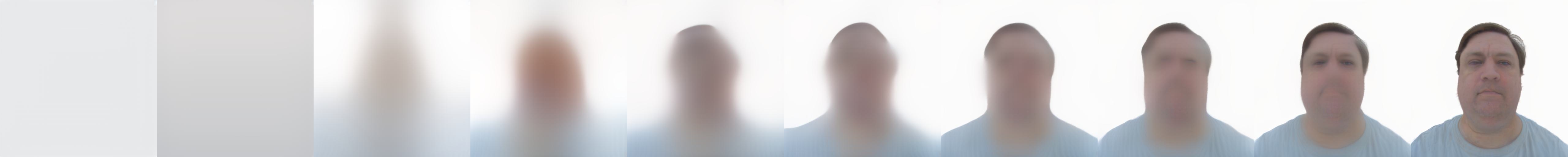}
    \end{minipage}
  \end{subfigure}
  
  \begin{subfigure}{\textwidth}
    \begin{minipage}[c]{0.25\textwidth}
      \vspace*{-0.6cm}%
      \raggedleft
      \scriptsize 512x512, 12 scales
    \end{minipage}%
    \hspace{0.5em}%
    \begin{minipage}[c]{0.75\textwidth}
      \sbox{\pnthirdimg}{\includegraphics[height=0.68cm]{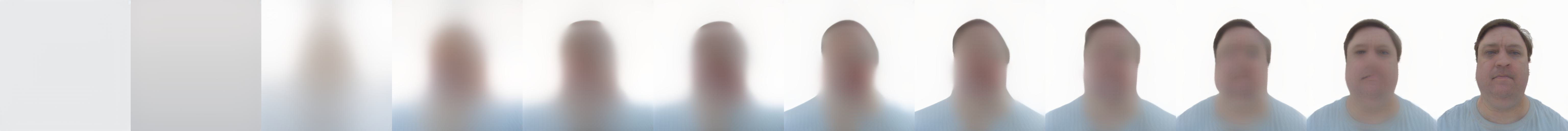}}
      \usebox{\pnthirdimg}
      \setlength{\pncolwidth}{\dimexpr\wd\pnthirdimg/12\relax}
      \\[2pt]
      \begingroup
      \setlength{\tabcolsep}{0pt}
      \begin{tabular}{@{}*{12}{c}@{}}
        \makebox[\pncolwidth][c]{\scriptsize $k_1$} & \makebox[\pncolwidth][c]{\scriptsize $k_2$} & \makebox[\pncolwidth][c]{\scriptsize $k_3$} & \makebox[\pncolwidth][c]{\scriptsize $k_4$} & \makebox[\pncolwidth][c]{\scriptsize $k_5$} & \makebox[\pncolwidth][c]{\scriptsize $k_6$} & \makebox[\pncolwidth][c]{\scriptsize $k_7$} & \makebox[\pncolwidth][c]{\scriptsize $k_8$} & \makebox[\pncolwidth][c]{\scriptsize $k_9$} & \makebox[\pncolwidth][c]{\scriptsize $k_{10}$} & \makebox[\pncolwidth][c]{\scriptsize $k_{11}$} & \makebox[\pncolwidth][c]{\scriptsize $k_{12}$} \\
      \end{tabular}
      \endgroup
    \end{minipage}
  \end{subfigure}
  \vspace{-0.2cm}
  \caption{Multi-scale VQ-VAE outputs for different resolutions and scale configurations. Notice that the small details appear at the last 2-3 scales, regardless of the resolution and the number of scales.}
  \label{fig:vqvae-multiscale}
\end{figure}

\subsection{Next-scale transformer for multi-view synthesis}
In the VAR architecture the goal is to shift the next-token paradigm of traditional transformers~\cite{attentionisallyouneed} for image generation~\cite{VQGAN}, which linearizes images through raster-scanning and generates them by predicting one token at a time, to a setting where the model generates multiple tokens at once, corresponding to an entire image scale. This is done multiple times, each corresponding to a different image scale, in order from lowest to highest detail. In its novel-view synthesis formulation, named ArchonView~\cite{archonview}, given an image conditioning $x$ and a target camera transformation $(R, T)$, the autoregressive likelihood of generating the set of multi-scale latent feature maps $\{r_1, r_2, \ldots, r_K\}$ is modeled as:
\begin{equation}
    g(x, R, T) = p_\theta(r_1, r_2, \ldots, r_K | x, R, T) = \prod_{k=1}^{K} p_\theta(r_k | r_{<k}, x, R, T)
\end{equation}
where $r_{<k} = \{r_1, r_2, \ldots, r_{k-1}\}$ are the previously generated scales, and $\theta$ are the learnable parameters of the transformer. The generation process starts from the lowest resolution scale $r_1$ and progressively generates higher resolution scales $r_2, r_3, \ldots, r_K$, conditioning each scale on the previously generated ones. The final tokens are predicted through a linear classification head as logits over the codebook entries.\\
We extend this framework to handle multiple input views and output views simultaneusly, as shown in Fig.~\ref{fig:pipeline}. Given N conditioning images $X = \{x_1, x_2, \allowbreak \ldots, x_N\}$ and M target camera transformations with respect to the first image $x_1$ $(R, T) = \{(R_1, T_1), (R_2, T_2), \ldots, (R_M, T_M)\}$, one per output view, the autoregressive likelihood of generating the set of multi-scale latent feature maps $\{r_{1,1}, \allowbreak r_{1,2}, \ldots, r_{1,M}, \ldots, r_{K,M}\}$ of $M$ output images is modeled as:

\begin{dmath} 
    g(X, R, T) = p_\theta(r_{1,1}, r_{1,2}, \ldots, r_{1,M}, \ldots, r_{K,M} | X, R, T) = \prod_{k=1}^{K} p_\theta(r_{k,1}, \ldots, r_{k,M} | r_{<k, 1}, \ldots, r_{<k, M}, X, R, T)
\end{dmath}
Notice that this formulation generates all output views at each scale before proceeding to the next scale, allowing the model to capture correlations between different views at multiple levels of detail, with the attention mask adjusted accordingly to allow cross-view attention within each scale, and complete availability of previous scales across all views.

\subsection{Input conditioning}
The input conditioning comprises both a global and a local conditioning. In~\cite{archonview}, for a single input image $x$ the global conditioning is represented by a posed CLIP~\cite{CLIP} embedding~\cite{zero1to3}:
\begin{equation}
    \tau(x, R, T) = W(W_i(\text{CLIP}(x)\oplus[\theta, sin\phi, cos\phi, r]) + b_i) + b
\end{equation}
where $W_i, b_i, W, b$ are learnable parameters of two linear layers, $\oplus$ is the concatenation operator, and $(\theta, \phi, r)$ are the spherical coordinates of the target camera transformation $(R, T)$. The first linear layer embeds the camera transformation into the CLIP features, while the second layer maps the resulting vector to the transformer embedding dimension, which is set as $w = 64d$, with $d$ the number of transformer blocks. The global conditioning is used both as the SOS token and in the AdaLN~\cite{adaln} layers inside transformer blocks. As in~\cite{archonview} and contrary to~\cite{VAR}, we empirically decide to avoid the use of AdaLN before the final classification head. To extend this to multiple output views, we compute a different SOS token for each view, using its view-specific $(R_m,T_m)$ while the AdaLN layers share the same global conditioning computed from the first output pose. For multiple input views, the global conditioning is computed only from the first one, which is considered the main view (near-frontal) and is the one used to define the target camera transformations.\\
The local conditioning, instead, consists of the multi-scale latent feature maps of the input image $x$ $\{i_1, i_2, \ldots, i_K\}$, obtained by encoding them with the frozen VQ-VAE encoder. When dealing with multiple input views $X = \{x_1, x_2, \ldots, x_N\}$, we concatenate their features together, obtaining $\{i_{1,1}, i_{2,1}, \ldots, i_{K,1}, \ldots, i_{K,N}\}$.

\subsection{Training strategy}
\label{sec:training_strategy}
Training a VAR architecture from scratch requires a large amount of data and computational resources. In~\cite{archonview}, the model is trained on the large-scale Objaverse~\cite{objaverse} dataset, which contains around 800k 3D objects from various categories, and it requires several days of training on hundreds of GPUs, at resolution $256 \times 256$ pixels. Training such a model specifically for photorealistic human heads is challenging both due to the lack of large-scale datasets and the need to generate higher-resolution images to capture details. Training a model from scratch at resolution $512 \times 512$ pixels would require even more data and computational resources, making it impractical. Fine-tuning a VAR model pre-trained at a different resolution or for a different purpose is also not straighforward, as the available high-quality human head data is not sufficient for the adaptation, as shown in~\cref{tab:ablation}. 
To overcome these challenges, we propose a three-stage training strategy, which allows us to effectively adapt a pre-trained VAR-derived model to our specific task by increasing its resolution and supporting multiple views. 

\paragraph{First stage.} We adapt the pre-trained ArchonView~\cite{archonview} model, which is trained for novel-view synthesis on objects, to work at a higher resolution of 512$\times$512 pixels. To do this we can either modify the architecture to support additional feature scales, corresponding to increasingily higher resolutions, or we can keep the same number of scales and increase the size of each scale instead. Increasing the number of scales makes intuitive sense, as it could leverage a model that already works at 256 by upscaling it at 512 using a few additional scales, but we argue that this is not optimal. In Fig.~\ref{fig:vqvae-multiscale} we show the scales of the VQ-VAE encoder for a sample image at resolution 256$\times$256 pixels (first row) and at resolution 512$\times$512 pixels using the same number of scales (second row) or using additional scales (third row). We observe that in every case the details appear progressively, but most of the facial features appear in the last two scales. This suggests that adding more scales would require a substantial change in the behavior of the transformer, which has learned to converge with a set number of scales. Therefore, we decide to keep the same number of scales and increase the size of each scale instead. In this first stage we train the model for general-purpose novel-view synthesis by using a proprietary large scale object dataset (SS3D, see \cref{sec:datasets}), selecting random input and target views. We validate this approach and the scale choice in~\cref{tab:ablation}. 

\paragraph{Second stage.} In this stage, we fine-tune the model to specialize it for photorealistic human head synthesis from frontal input views. We use the PSGS~\cite{psgs} dataset, with around 2.9K training subjects, and we train the model to generate random turntable views. We specialize two separate models, one that deals with a single frontal input view, and one that deals with two additional side views at approximately $\pm 90$ degrees yaw. These additional views are loosely posed: their ground truths are generated with 32 camera and head pose variations per subject, mimicking real-world selfie captures (see Fig.~\ref{fig:pipeline}). In this stage, we also fine-tune the VQ-VAE weights from VAR~\cite{VAR}, which were trained on the OpenImages~\cite{openimages} dataset at a resolution of $256 \times 256$ pixels. To adapt them to our task, we finetune them on the PSGS~\cite{psgs} dataset of human heads at resolution 512$\times$512. Notice that, at this stage, the models are not constrained to output canonical views, and can output any turntable view.

\paragraph{Third stage.} Finally, we further specialize the models to symultaneously predict the 6 canonical views needed for the subsequent 3D reconstruction stage, using the same data as in the second stage. This allows the model to better capture correlations between the different views, improving the overall quality and consistency of the generated images. To achieve this, we empirically determined that classifier-free guidance is not beneficial (see \cref{fig:adapting_and_cfg}, right), so we do not employ it in this training stage. We conjecture that the limited amount of training data, alongside with the model being specialized from a general-purpose one, contribute to this behavior.

  \section{Experiments}

\subsection{Datasets}
\label{sec:datasets}
\paragraph{SS3D.} A proprietary large scale object dataset, containing around 2M unique textured objects. They have been rendered at resolution of 512x512 from 24 random cameras sampled on a sphere and using area lights.
\paragraph{PSGS~\cite{psgs}.} A proprietary dataset obtained by capturing calibrated multi-view images of real individuals from a dome capture setup, and fitting a high-fidelity gaussian avatar to each of them, which can then be rendered with arbitrary camera extrinsics and intrinsics with head pose variations. The dataset is more realistic but smaller than existing synthetic datasets~\cite{facelift,volume}, as it consists of 3.2K subjects, of which 288 are used for validation. We render the subjects from 32 turntable views at a fixed distance. We additionally render 32 variants per subject, each with a randomly perturbed camera pose and head pose, to use as loosely-posed training inputs for our multi-input model.
\paragraph{Ava-256~\cite{ava256}.} A publicly released multi-view human-head dataset comprising 256 subjects, with 80 high-resolution dome camera views per subject.

\subsection{Metrics}
Across our experiments we employ the following metrics, which are standard in the field~\cite{facelift, volume}: Peak Signal to Noise Ratio (PSNR), to measure the overall quality; Structural Similarity Index Measure (SSIM), to measure the perceived similarity; Learned Perceptual Image Patch Similarity (LPIPS)~\cite{lpips}, with the VGG~\cite{vgg} backbone, to measure the distance between images in feature space; DreamSim (DS)~\cite{dreamsim1,dreamsim2}, which measures image distance by focusing on mid-level features, fine-tuned on human perceptual judgements; ArcFace~\cite{arcface}, measuring human face-centric identity shift as cosine similarity in feature space. Since it requires face detection to be measured, if a face is detected in the predicted image but not in the ground truth or viceversa we define the score as 0 (worst), whereas if a face is not detected in any of them, we define the score as 1 (best).

\subsection{Baselines}
There exist numerous methods that tackle similar problems --- see~\cref{sec:related}. To the best of our knowledge, the most recent and accurate baselines for 3D face reconstruction without requiring camera poses are FaceLift~\cite{facelift} and VoluMe~\cite{volume}.\\
FaceLift~\cite{facelift} is divided in two models: the first one is a diffusion-based novel view synthesis model that predicts 6 canonical views from a single input image, akin to our multi-output model, and is trained from Stable Diffusion V2-1-unCLIP ~\cite{stablediffusion}; the second one is a GS-LRM model that lifts the 6 views from the previous model into pixel-aligned 3D gaussians, which can then be rendered from any view. We refer to the full pipeline as \textit{FaceLift}, and we refer to the first half of the pipeline as \textit{FaceLift-NVS}. Moreover, since the data used to train the model is not publicly available, we ensure a fair comparison with our model by also re-training FaceLift-NVS with our PSGS data directly from the Stable Diffusion checkpoint. We refer to this model as \textit{FaceLift-NVS (PSGS)}.\\
VoluMe~\cite{volume} is single-stage pipeline that directly predicts pixel-aligned 3D gaussians from a single input image. Neither the code nor the data used for VoluMe are publicly available at the time of writing, making a direct comparison impossible. However VoluMe is based on Splatter Image~\cite{splatterimage}, which is shown in~\cite{volume} to produce similar results when trained on the same data. Thus, we train it on our PSGS data and employ it as our baseline, named \textit{SI (PSGS)}. Due to VRAM requirements, it has been trained at resolution 256x256. The model is trained in two stages, we include results from both stages. If it is not specified, it is assumed to be the final model.\\
For a fair comparison we use 24 transformer blocks in our models, for a total of approximately 1B parameters, which is comparable to the size of Facelift-NVS. Additional training details are available in the supplementary material.


\subsection{Experimental results}
\begin{table*}[h!]
\centering
\caption{Quantitative evaluation on PSGS~\cite{psgs} validation set for NVS models (6 canonical views). Best in bold, second best in italics.}
\begin{tabularx}{\textwidth}{l|*{5}{>{\centering\arraybackslash}X}}
& PSNR $\uparrow$ & SSIM $\uparrow$ & LPIPS $\downarrow$ & DS $\downarrow$ & ArcFace $\uparrow$ \\
\hline
SI (PSGS) - I stage~\cite{splatterimage} & 20.15 & 0.8640 & 0.2612 & 0.02606 & 0.6761 \\

SI (PSGS) - II stage~\cite{splatterimage} & 20.16 & 0.8614 & 0.2351 & 0.02297 & \textbf{0.7403} \\

FaceLift-NVS~\cite{facelift} & 17.76 & 0.7942 & 0.2785 & 0.04531 & 0.6554 \\

FaceLift-NVS (PSGS)\cite{facelift} & 16.75 & 0.7900 & 0.3206 & 0.02262 & 0.6680 \\

\textit{Ours} & 22.29 & \textit{0.8806} & \textit{0.1905} & 0.02191 & 0.6944 \\

\textit{Ours MO} & \textit{22.40} & 0.8752 & 0.1927 & \textit{0.01729} & 0.7069 \\

\textit{Ours MI-MO} & \textbf{23.13} & \textbf{0.8816} & \textbf{0.1762} & \textbf{0.01379} & \textit{0.7206} \\

\end{tabularx}
\label{tab:psgs_nvs_bench}
\end{table*}

\begin{figure*}[ht]
    \centering
    \begin{minipage}[t]{0.48\textwidth}
        \begin{minipage}[c]{0.18\linewidth}
            \centering
            \scriptsize Inputs
        \end{minipage}%
        \hspace{0.01\linewidth}
        \begin{minipage}[c]{0.88\linewidth}
            \includegraphics[width=0.25\linewidth]{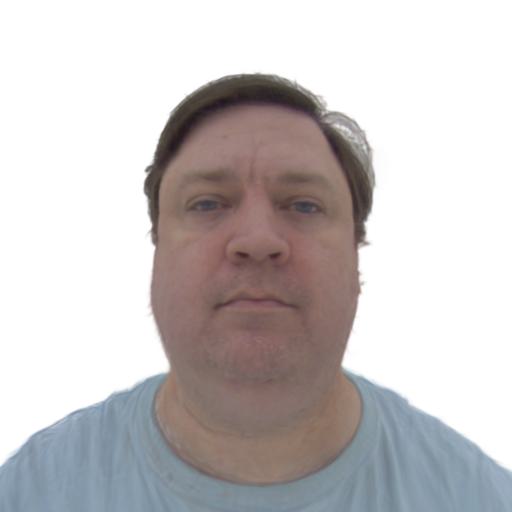}
            \hspace{0.01\linewidth}
            \vrule width 1pt height 0.18\linewidth
            \hspace{0.01\linewidth}
            \includegraphics[width=0.25\linewidth]{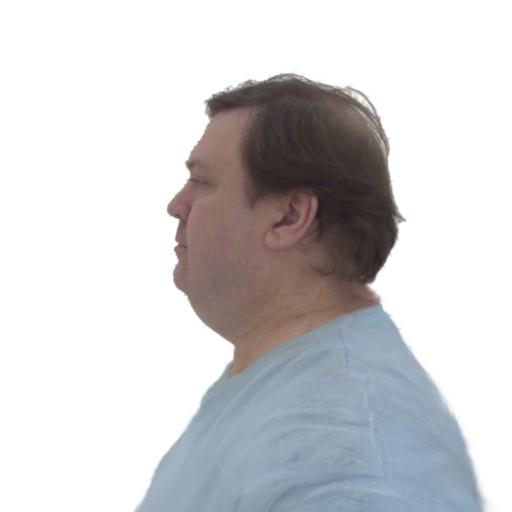}
            \hspace{0.01\linewidth}
            \includegraphics[width=0.25\linewidth]{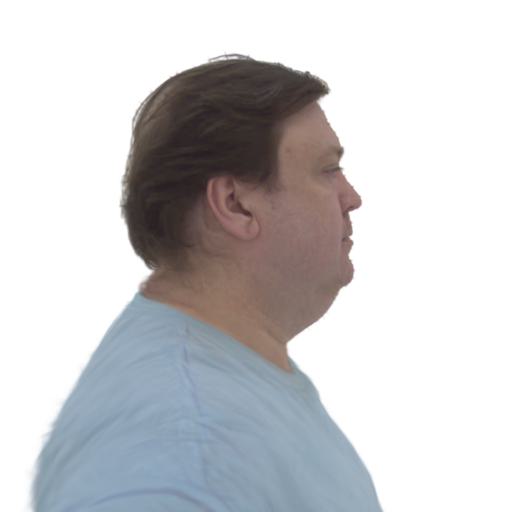}
        \end{minipage}

        \par\nointerlineskip

                \begin{minipage}[c]{0.18\linewidth}
            \centering
            \scriptsize GT
    \end{minipage}%
    \hspace{0.01\linewidth}
    \begin{minipage}[c]{0.8\linewidth}
        \centering
        \includegraphics[width=1\linewidth]{figs/nvs1/gt.jpg}
    \end{minipage}
    \par\nointerlineskip
    \begin{minipage}[c]{0.18\linewidth}
        \centering
        \scriptsize SI (PSGS) I
    \end{minipage}%
    \hspace{0.01\linewidth}
    \begin{minipage}[c]{0.8\linewidth}
        \centering
        \includegraphics[width=1\linewidth]{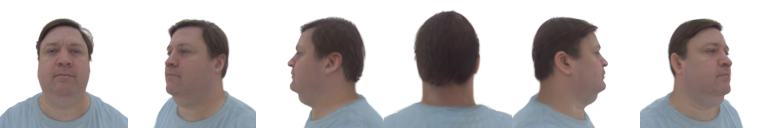}
    \end{minipage}
    \par\nointerlineskip
    \begin{minipage}[c]{0.18\linewidth}
        \centering
        \scriptsize SI (PSGS) II
    \end{minipage}%
    \hspace{0.01\linewidth}
    \begin{minipage}[c]{0.8\linewidth}
        \centering
        \includegraphics[width=1\linewidth]{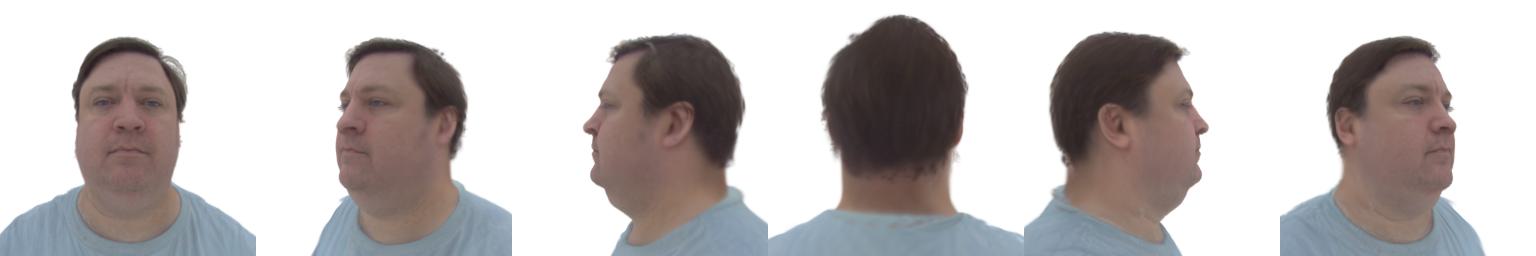}
    \end{minipage}
    \par\nointerlineskip
    \begin{minipage}[c]{0.18\linewidth}
        \centering
        \scriptsize FaceLift-NVS
    \end{minipage}%
    \hspace{0.01\linewidth}
    \begin{minipage}[c]{0.8\linewidth}
        \centering
        \includegraphics[width=1\linewidth]{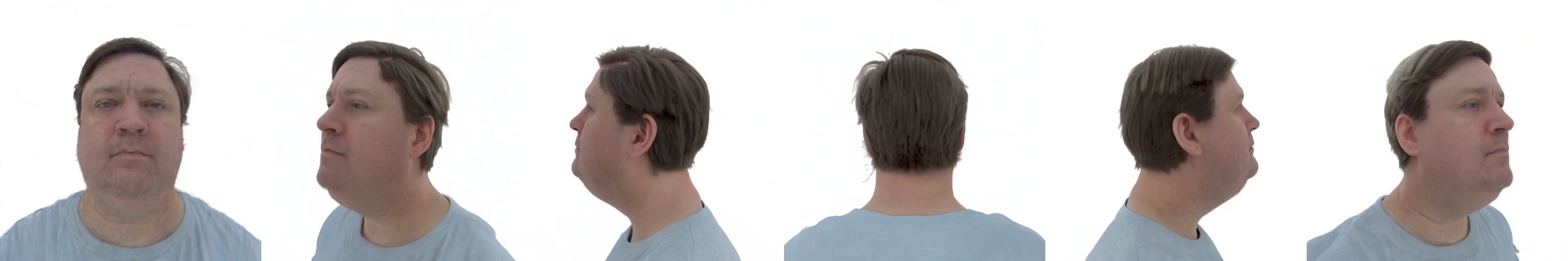}
    \end{minipage}
    \par\nointerlineskip
    \begin{minipage}[c]{0.18\linewidth}
        \centering
        \scriptsize FaceLift-NVS (PSGS)
    \end{minipage}%
    \hspace{0.01\linewidth}
    \begin{minipage}[c]{0.8\linewidth}
        \centering
        \includegraphics[width=1\linewidth]{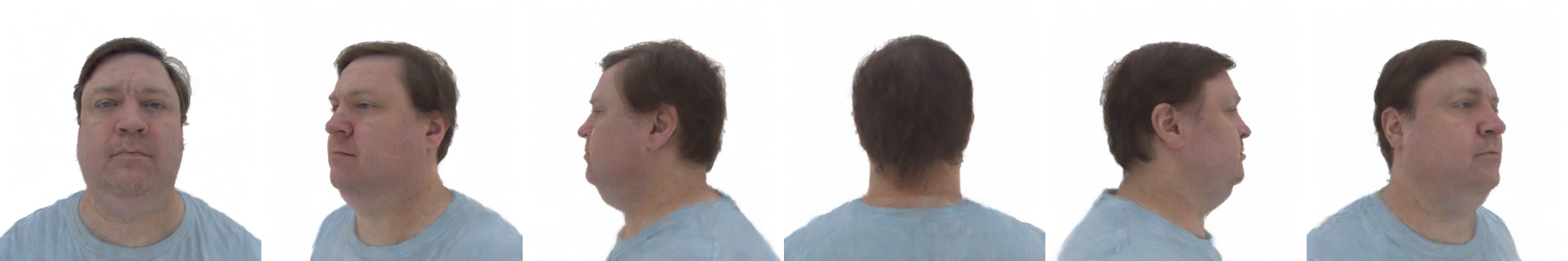}
    \end{minipage}
    \par\nointerlineskip
    \begin{minipage}[c]{0.18\linewidth}
        \centering
        \scriptsize \textit{Ours}
    \end{minipage}%
    \hspace{0.01\linewidth}
    \begin{minipage}[c]{0.8\linewidth}
        \centering
        \includegraphics[width=1\linewidth]{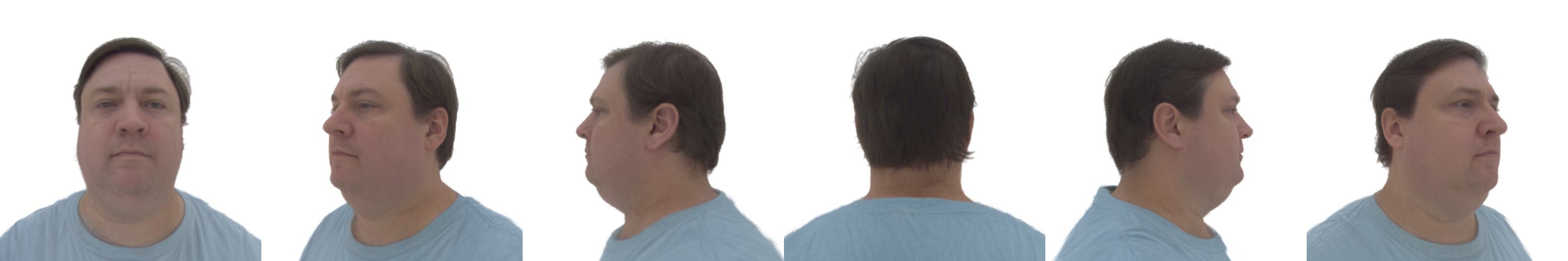}
    \end{minipage}
    \par\nointerlineskip
    \begin{minipage}[c]{0.18\linewidth}
        \centering
        \scriptsize \textit{Ours MO}
    \end{minipage}%
    \hspace{0.01\linewidth}
    \begin{minipage}[c]{0.8\linewidth}
        \centering
        \includegraphics[width=1\linewidth]{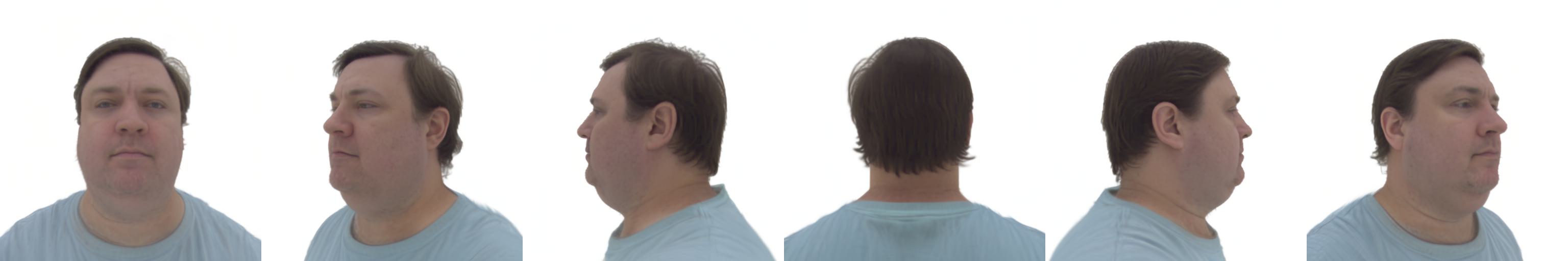}
    \end{minipage}
    \par\nointerlineskip
    \begin{minipage}[c]{0.18\linewidth}
        \centering
        \scriptsize \textit{Ours MI-MO}
    \end{minipage}%
    \hspace{0.01\linewidth}
    \begin{minipage}[c]{0.8\linewidth}
        \centering
        \includegraphics[width=1\linewidth]{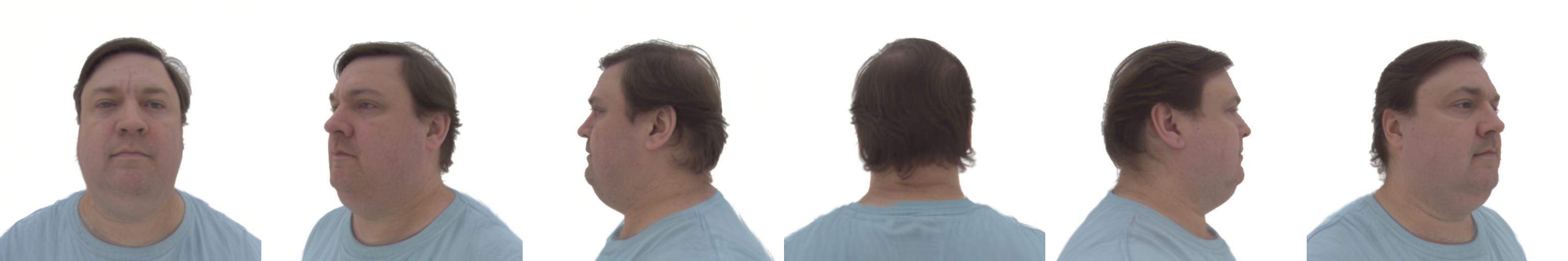}
    \end{minipage}

    \end{minipage}%
    \hspace{0.02\textwidth}
    \begin{minipage}[t]{0.48\textwidth}
        \begin{minipage}[c]{0.88\linewidth}
            \centering
            \hspace{0.09\linewidth}
            \includegraphics[width=0.25\linewidth]{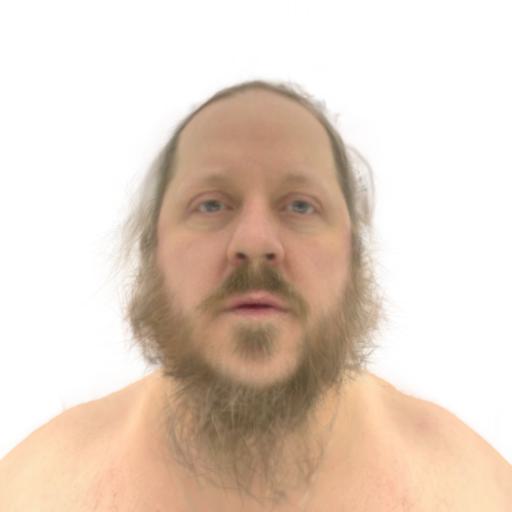}
            \hspace{0.01\linewidth}
            \vrule width 1pt height 0.18\linewidth
            \hspace{0.01\linewidth}
            \includegraphics[width=0.25\linewidth]{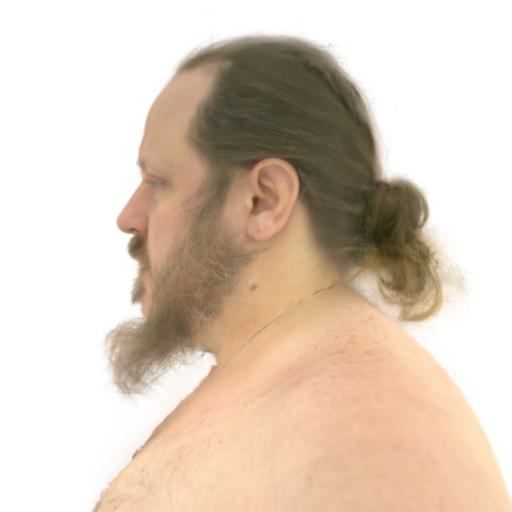}
            \hspace{0.01\linewidth}
            \includegraphics[width=0.25\linewidth]{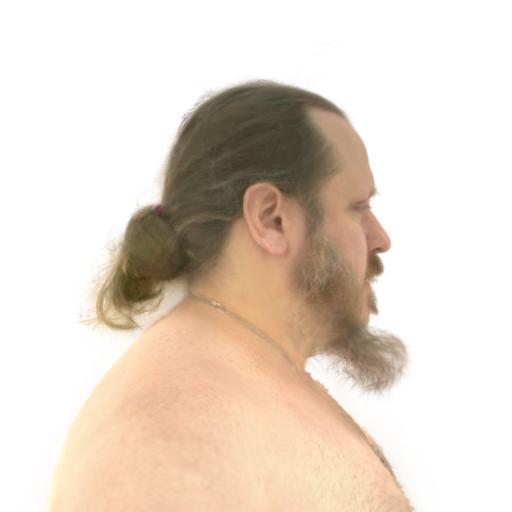}
        \end{minipage}
        \par\nointerlineskip
        \begin{minipage}[c]{\linewidth}
            \centering
            \includegraphics[width=0.8\linewidth]{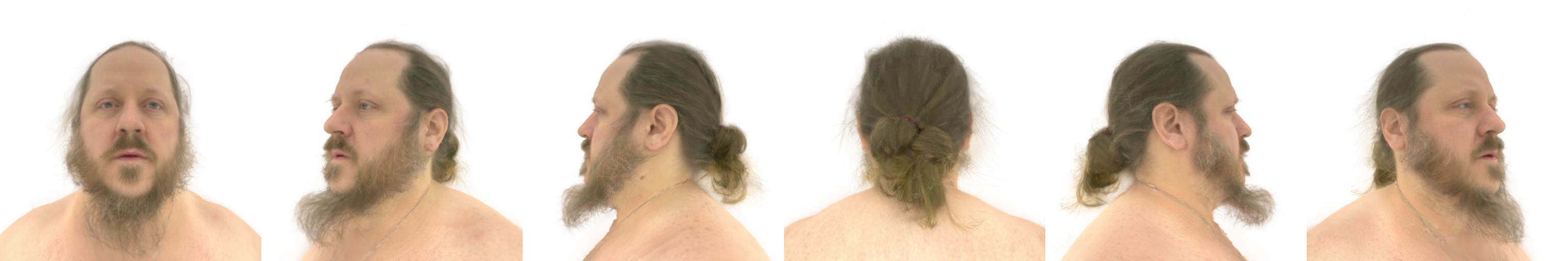}
        \end{minipage}
        \par\nointerlineskip
        \begin{minipage}[c]{\linewidth}
            \centering
            \includegraphics[width=0.8\linewidth]{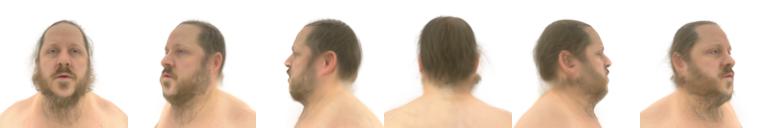}
        \end{minipage}
        \par\nointerlineskip
        \begin{minipage}[c]{\linewidth}
            \centering
            \includegraphics[width=0.8\linewidth]{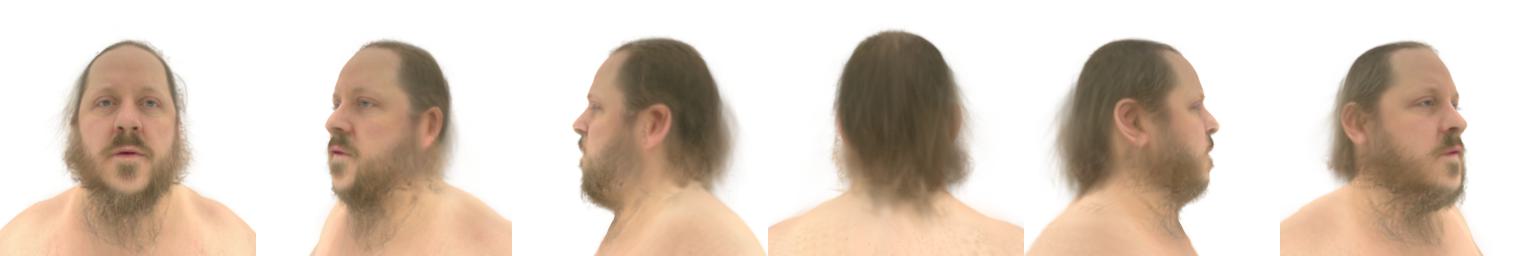}
        \end{minipage}
        \par\nointerlineskip
        \begin{minipage}[c]{\linewidth}
            \centering
            \includegraphics[width=0.8\linewidth]{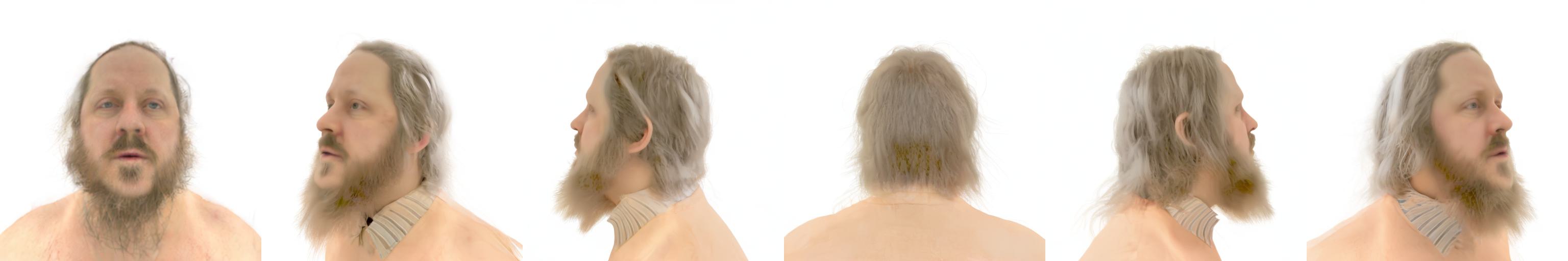}
        \end{minipage}
        \par\nointerlineskip
        \begin{minipage}[c]{\linewidth}
            \centering
            \includegraphics[width=0.8\linewidth]{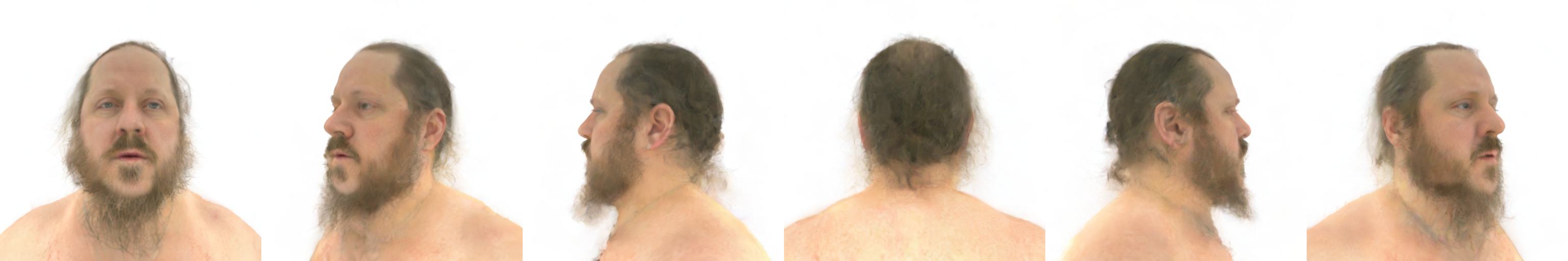}
        \end{minipage}
        \par\nointerlineskip
        \begin{minipage}[c]{\linewidth}
            \centering
            \includegraphics[width=0.8\linewidth]{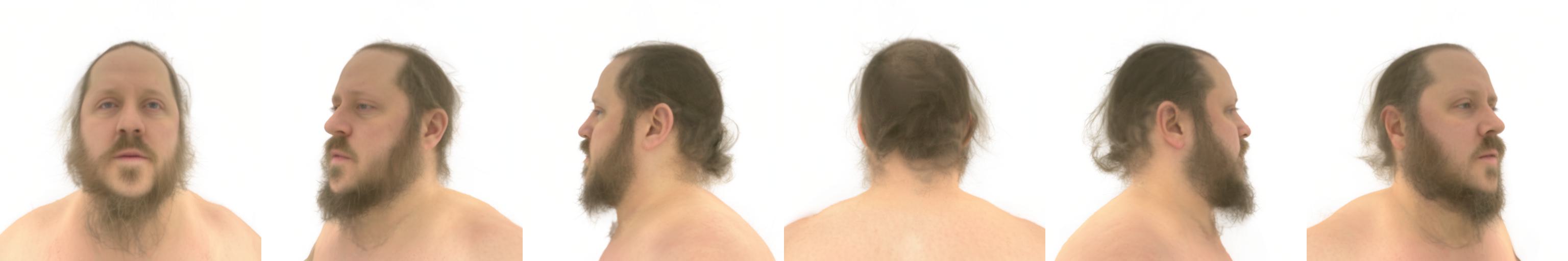}
        \end{minipage}
        \par\nointerlineskip
        \begin{minipage}[c]{\linewidth}
            \centering
            \includegraphics[width=0.8\linewidth]{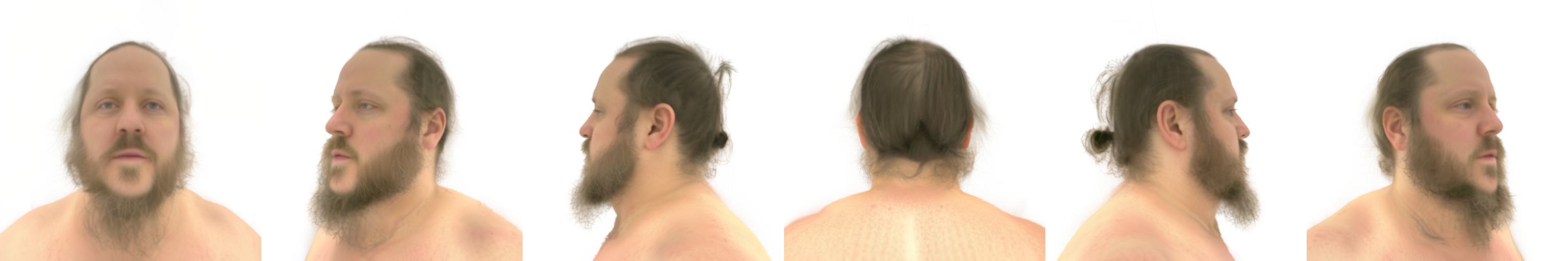}
        \end{minipage}
        \par\nointerlineskip
        \begin{minipage}[c]{\linewidth}
            \centering
            \includegraphics[width=0.8\linewidth]{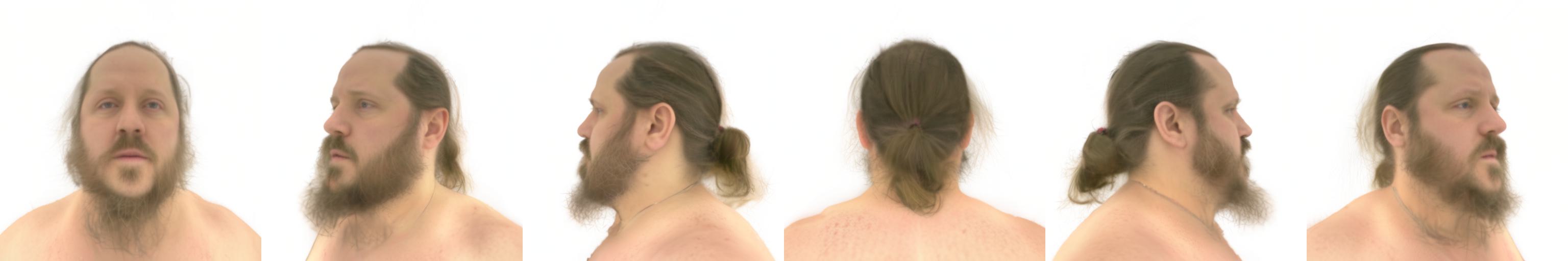}
        \end{minipage}

    \end{minipage}
    
    \caption{Qualitative results on PSGS~\cite{psgs} validation set on the 6 canonical views.}
    \label{fig:nvs1-2}
\end{figure*}

We show quantitative and qualitative results for the novel view synthesis stage of the pipeline on the PSGS dataset in~\cref{tab:psgs_nvs_bench} and~\cref{fig:nvs1-2}, using the 6 canonical output views defined by FaceLift~\cite{facelift}. Our approach outperforms the baselines across most metrics, with the exception of ArcFace, where Splatter Image (PSGS) performs sligthly better. Hoerver, while it is able to achieve good results in terms of identity preservation in near-frontal views, it struggles to produce high-quality back and side views, because it is not designed to do so, resulting in blurry images, as visible in \cref{fig:nvs1-2}. It also struggles to correctly center the head, resulting in slight misalignments in the side views. Our method is designed to produce high-quality results across all views, and to do so with correct alignment for use in downstream 3D reconstruction pipelines. FaceLift-NVS, instead, tends to produce very detailed but artificial-looking results, likely due to its reliance on a synthetic training dataset. When it is trained on our fewer but more realistic data the results become less detailed but more realistic. We also notice that, compared to our method, FaceLift-NVS (PSGS) tends to produce small visual artifacts, hindering its overall quality. We conjecture that this is due to the small amount of training data, which might not be sufficient to easily train a large diffusion model. Our model, instead, is able to better leverage the training data and correctly converges even with small amounts of data, thanks to the training strategy we propose. Compared to our single-output model, our multi-output model (MO) is able to further improve the metrics thanks to the cross-attention between the different views during generation. This is also visible in \cref{fig:nvs1-2} (right), where \textit{Ours MO} is more cross-view consistent in terms of beard and hair style compared to \textit{Ours}. Moreover, the multi-input multi-output model (MI-MO) is able to further improve the results by leveraging the additional loosely-posed input views, producing very realistic and consistent non-frontal views. It is also worth noting that, while FaceLift requires around 5 seconds to generate the 6 views on our setup, our method is able to generate the same views in around 1.5 seconds. The fastest method remains Splatter Image which, while not designed to render non-frontal views, can run in real time.

We compare the baselines on the same dataset using 32 turntable views in~\cref{tab:psgs_3d_bench} and~\cref{fig:turntable1}. While Splatter Image can naturally render these views, for the other methods we use the pipelines described in \cref{sec:method}. The results are consistent with the ones obtained on the 6 canonical views, with our method outperforming the baselines across the metrics. Notice that, similarly to Splatter Image, our single-input single-output model is not trained to produce only the 6 canonical views, and thus can be used to render any target view of the subject without an additional step (see \cref{fig:turntable1}, row 7). Compared to Splatter Image, this approach produces much more detailed results, including non-frontal views, but with inconsistencies across views, as it is not designed to handle them. When consistency is needed, the full pipeline should be employed.

We also show similar results on the Ava-256~\cite{ava256} dataset in~\cref{tab:ava256_bench} and~\cref{fig:ava1}, using the 10 subjects and 11 views defined by FaceLift~\cite{facelift}. These results have been obtained following the experimental description of FaceLift~\cite{facelift} and VoluMe~\cite{volume}. Due to camera-pose definitions in this dataset, the original formulation of this experiment required manual aligment. Since the experimental code has not been released by neither~\cite{facelift} nor~\cite{volume}, it is impossible to directly compare results, as well as to ensure unbiased alignment. As a consequence, this dataset remains challenging to work with in this specific scenario, and we believe its results to be less reliable compared to previous experiments. However, to partially overcome this limitation we implement an automatic background removal and alignment procedure based on Rembg~\cite{rembg}, and apply it to all methods. The final results are similar to our previous benchmarks, with the most notable difference being the improved performance of FaceLift in terms of identity preservation, likely due to its larger training dataset, compared to when it is trained with PSGS data. The results on the other metrics are largely unchanged and, with the same training data, our method fully outperforms FaceLift.

We additionally refer the reader to our supplementary material for qualitative results on full body data.

\begin{figure*}[ht]
    \centering
        \begin{minipage}[c]{0.22\linewidth}
            \centering
            \small Inputs
        \end{minipage}%
        \hspace{0.01\linewidth}
        \begin{minipage}[c]{0.76\linewidth}
            \includegraphics[width=0.14\linewidth]{figs/turntable1/input1.jpg}
            \hspace{0.0125\linewidth}
            \vrule width 1pt height 0.1\linewidth
            \hspace{0.0125\linewidth}
            \includegraphics[width=0.14\linewidth]{figs/turntable1/input2.jpg}
            \hspace{0.0125\linewidth}
            \includegraphics[width=0.14\linewidth]{figs/turntable1/input3.jpg}
        \end{minipage}

        \par\nointerlineskip

                \begin{minipage}[c]{0.22\linewidth}
        \centering
        \scriptsize GT
    \end{minipage}%
    \hspace{0.01\linewidth}
    \begin{minipage}[c]{0.76\linewidth}
        \centering
        \includegraphics[width=1\linewidth]{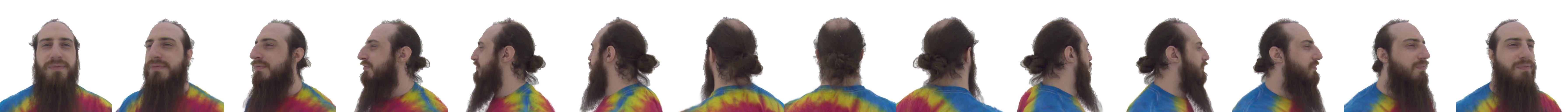}
    \end{minipage}
    \par\nointerlineskip
    \begin{minipage}[c]{0.22\linewidth}
        \centering
        \scriptsize SI (PSGS) I
    \end{minipage}%
    \hspace{0.01\linewidth}
    \begin{minipage}[c]{0.76\linewidth}
        \centering
        \includegraphics[width=1\linewidth]{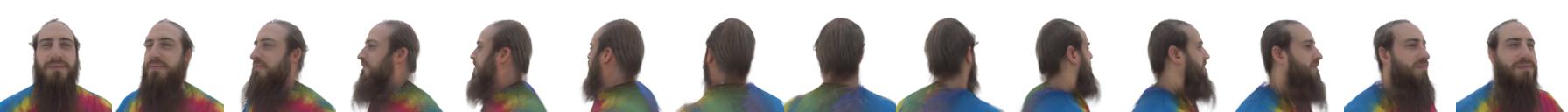}
    \end{minipage}
    \par\nointerlineskip
    \begin{minipage}[c]{0.22\linewidth}
        \centering
        \scriptsize SI (PSGS) II
    \end{minipage}%
    \hspace{0.01\linewidth}
    \begin{minipage}[c]{0.76\linewidth}
        \centering
        \includegraphics[width=1\linewidth]{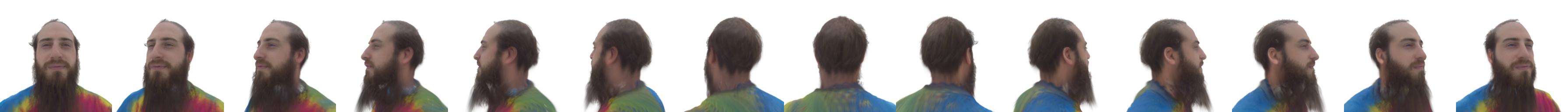}
    \end{minipage}
    \par\nointerlineskip
    \begin{minipage}[c]{0.22\linewidth}
        \centering
        \scriptsize FaceLift
    \end{minipage}%
    \hspace{0.01\linewidth}
    \begin{minipage}[c]{0.76\linewidth}
        \centering
        \includegraphics[width=1\linewidth]{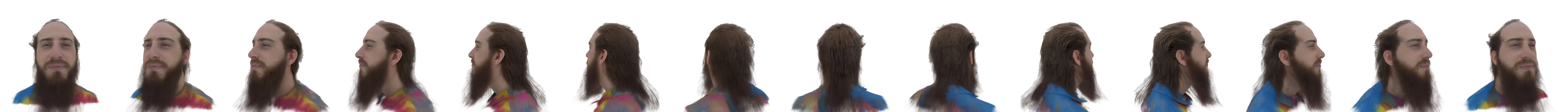}
    \end{minipage}
    \par\nointerlineskip
    \begin{minipage}[c]{0.22\linewidth}
        \centering
        \scriptsize FaceLift (PSGS)
    \end{minipage}%
    \hspace{0.01\linewidth}
    \begin{minipage}[c]{0.76\linewidth}
        \centering
        \includegraphics[width=1\linewidth]{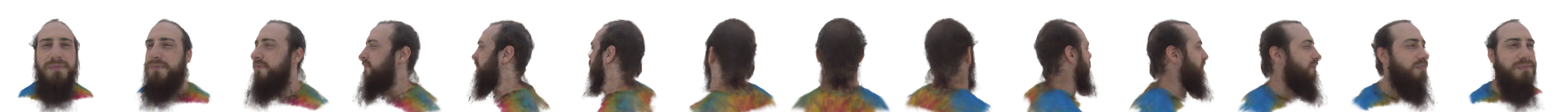}
    \end{minipage}
    \par\nointerlineskip
    \begin{minipage}[c]{0.22\linewidth}
        \centering
        \scriptsize \textit{Ours}
    \end{minipage}%
    \hspace{0.01\linewidth}
    \begin{minipage}[c]{0.76\linewidth}
        \centering
        \includegraphics[width=1\linewidth]{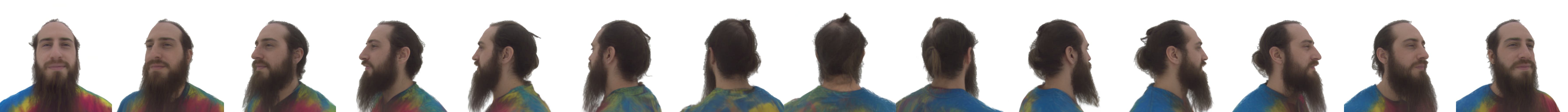}
    \end{minipage}
    \par\nointerlineskip
    \begin{minipage}[c]{0.22\linewidth}
        \centering
        \scriptsize \textit{Ours +~\cite{psgs}}
    \end{minipage}%
    \hspace{0.01\linewidth}
    \begin{minipage}[c]{0.76\linewidth}
        \centering
        \includegraphics[width=1\linewidth]{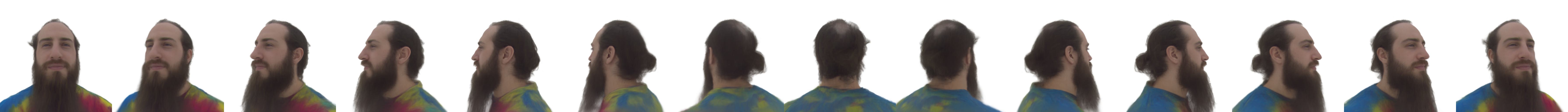}
    \end{minipage}
    \par\nointerlineskip
    \begin{minipage}[c]{0.22\linewidth}
        \centering
        \scriptsize \textit{Ours MO +~\cite{psgs}}
    \end{minipage}%
    \hspace{0.01\linewidth}
    \begin{minipage}[c]{0.76\linewidth}
        \centering
        \includegraphics[width=1\linewidth]{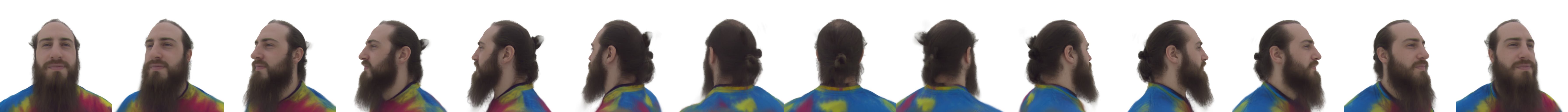}
    \end{minipage}
    \par\nointerlineskip
    \begin{minipage}[c]{0.22\linewidth}
        \centering
        \scriptsize \textit{Ours MI-MO +~\cite{psgs}}
    \end{minipage}%
    \hspace{0.01\linewidth}
    \begin{minipage}[c]{0.76\linewidth}
        \centering
        \includegraphics[width=1\linewidth]{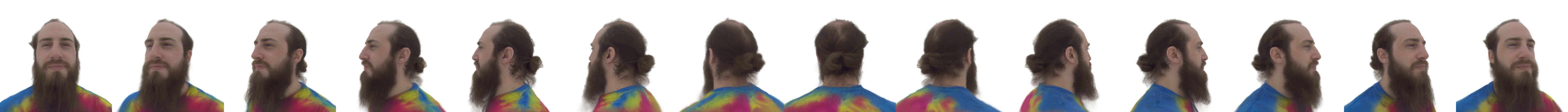}
    \end{minipage}
    \caption{Qualitative results on PSGS~\cite{psgs} validation set on the 32 turntable views.}
    \label{fig:turntable1}
\end{figure*}
\begin{table*}[h!]
\centering
\caption{Quantitative evaluation on PSGS~\cite{psgs} validation set of the full image-to-3D pipelines (32 turntable views). Best in bold, second best in italics.}
\begin{tabularx}{\textwidth}{l|*{5}{>{\centering\arraybackslash}X}}
& PSNR $\uparrow$ & SSIM $\uparrow$ & LPIPS $\downarrow$ & DS $\downarrow$ & ArcFace $\uparrow$ \\
\hline
SI (PSGS) - I stage~\cite{splatterimage} & 18.46 & 0.8548 & 0.2755 & 0.03257 & 0.7108 \\
SI (PSGS) - II stage~\cite{splatterimage} & 18.32 & 0.8488 & 0.2549 & 0.03184 & 0.7596 \\
FaceLift~\cite{facelift} & 13.28 & 0.7791 & 0.3103 & 0.1008 & 0.7121 \\
FaceLift (PSGS)~\cite{facelift} & 15.41 & 0.8377 & 0.2668 & 0.07339 & 0.7357 \\
\textit{Ours} +~\cite{psgs} & 20.78 & \textit{0.8805} & \textit{0.2248} & 0.02440 & 0.7520 \\
\textit{Ours MO} +~\cite{psgs} & \textit{20.82} & 0.8772 & 0.2249 & \textit{0.02319} & \textit{0.7608} \\
\textit{Ours MI-MO} +~\cite{psgs} & \textbf{21.91} & \textbf{0.8865} & \textbf{0.2110} & \textbf{0.01188} & \textbf{0.7737} \\
\end{tabularx}
\label{tab:psgs_3d_bench}
\end{table*}
\begin{table*}[h!]
\centering
\caption{Quantitative evaluation on AVA-256, 10 subjects, 11 views, as defined in~\cite{facelift}. Best in bold, second best in italics.}
\label{tab:ava256_bench}
\begin{tabularx}{\textwidth}{l|*{5}{>{\centering\arraybackslash}X}}
& PSNR $\uparrow$ & SSIM $\uparrow$ & LPIPS $\downarrow$ & DS $\downarrow$ & ArcFace $\uparrow$ \\
\hline
SI (PSGS)~\cite{splatterimage} & 13.06 & 0.7736 & 0.2911 & 0.06242 & \textit{0.6689} \\
FaceLift~\cite{facelift} & 14.49 & 0.8038 & 0.2546 & \textit{0.03482} & \textbf{0.7699} \\
FaceLift (PSGS)~\cite{facelift} & 14.51 & 0.8146 & 0.2489 & 0.03584 & 0.5837 \\
\textit{Ours} +~\cite{psgs} & \textit{15.34} & \textit{0.8201} & \textit{0.2357} & 0.03616 & 0.6275 \\
\textit{Ours MO} +~\cite{psgs} & 15.24 & 0.8197 & 0.2448 & \textbf{0.03385} & 0.6405 \\
\textit{Ours MI-MO} +~\cite{psgs} & \textbf{15.67} & \textbf{0.8209} & \textbf{0.2350} & 0.04187 & 0.6251 \\
\end{tabularx}
\vspace{-0.5cm}
\end{table*}

\begin{figure*}[ht]
    \centering
        \begin{minipage}[c]{0.22\linewidth}
            \centering
            \small Inputs
        \end{minipage}%
        \hspace{0.01\linewidth}
        \begin{minipage}[c]{0.76\linewidth}
            \includegraphics[width=0.13\linewidth]{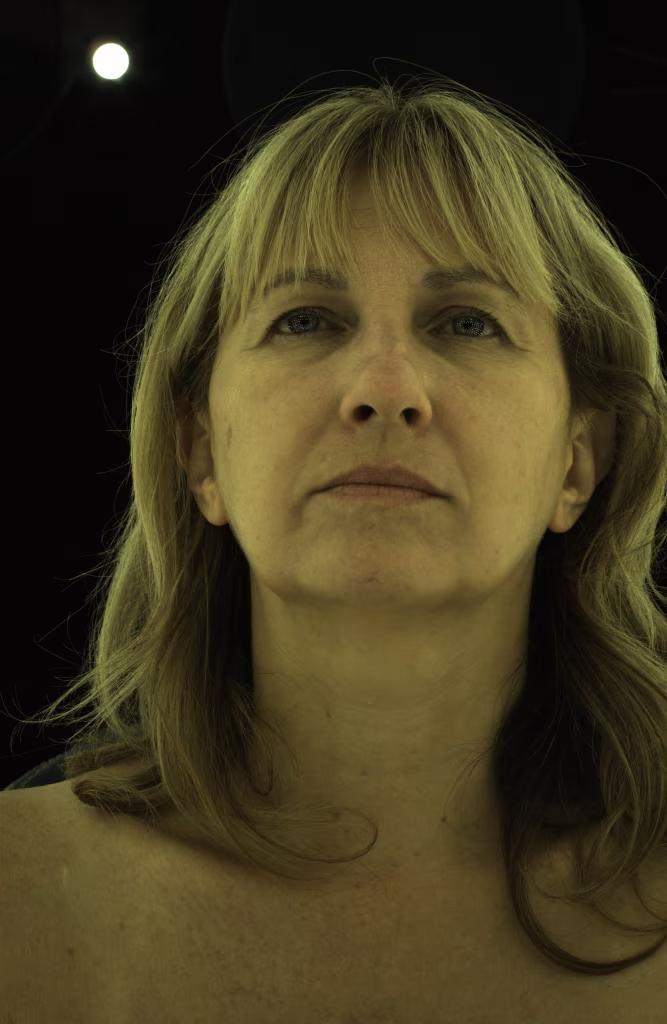}
            \hspace{0.01\linewidth}
            \vrule width 1pt height 0.2\linewidth
            \hspace{0.01\linewidth}
            \includegraphics[width=0.13\linewidth]{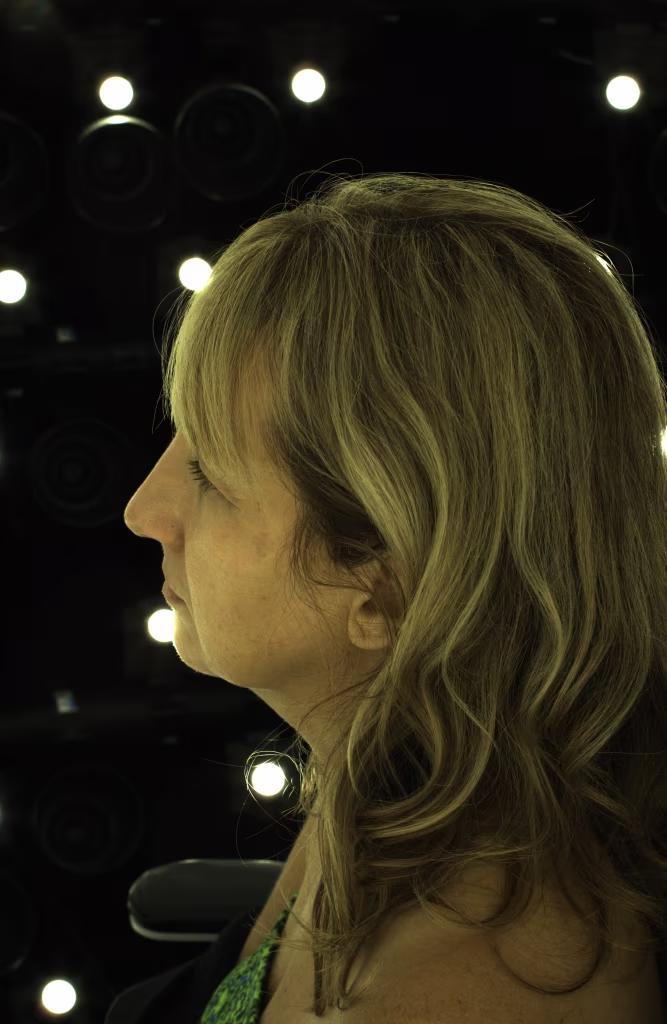}
            \hspace{0.01\linewidth}
            \includegraphics[width=0.13\linewidth]{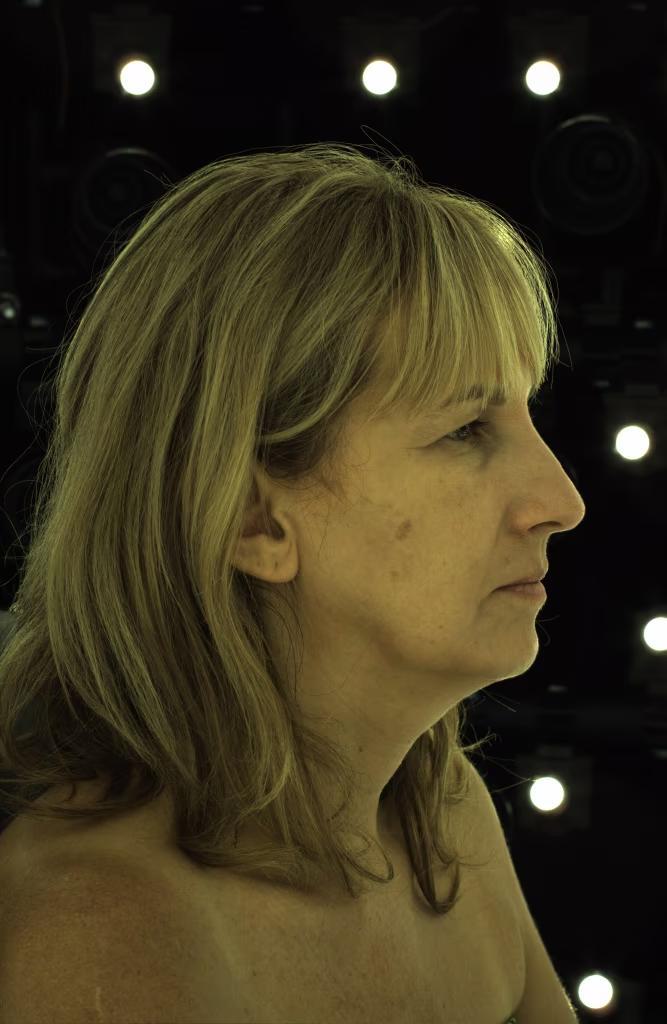}
        \end{minipage}

        \par\vspace{1mm}

                \begin{minipage}[c]{0.22\linewidth}
        \centering
        \scriptsize GT
    \end{minipage}%
    \hspace{0.01\linewidth}
    \begin{minipage}[c]{0.76\linewidth}
        \centering
        \includegraphics[width=1\linewidth]{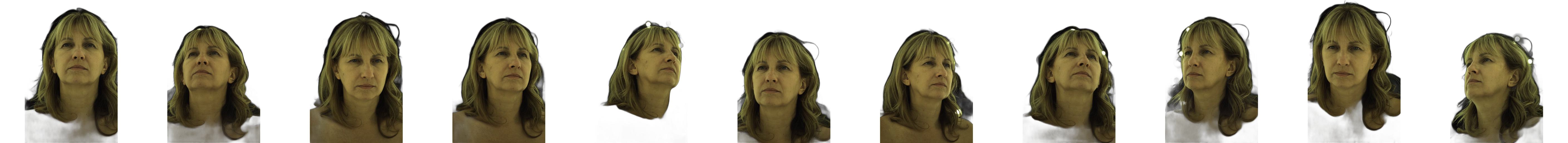}
    \end{minipage}
    \par\vspace{1mm}
    \begin{minipage}[c]{0.22\linewidth}
        \centering
        \scriptsize SI (PSGS)
    \end{minipage}%
    \hspace{0.01\linewidth}
    \begin{minipage}[c]{0.76\linewidth}
        \centering
        \includegraphics[width=1\linewidth]{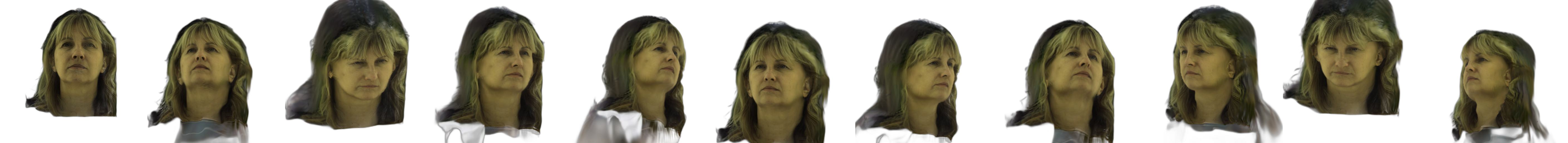}
    \end{minipage}
    \par\vspace{1mm}
    \begin{minipage}[c]{0.22\linewidth}
        \centering
        \scriptsize FaceLift
    \end{minipage}%
    \hspace{0.01\linewidth}
    \begin{minipage}[c]{0.76\linewidth}
        \centering
        \includegraphics[width=1\linewidth]{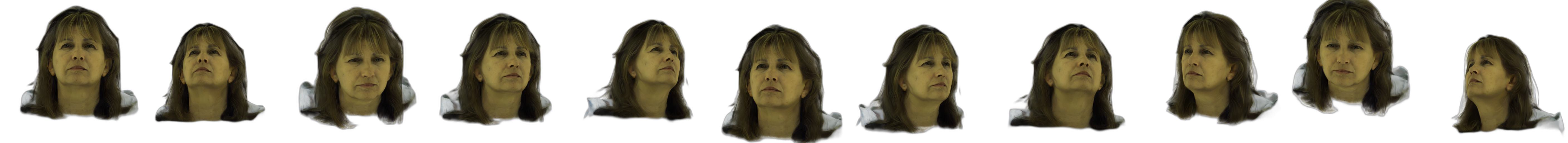}
    \end{minipage}
    \par\vspace{1mm}
    \begin{minipage}[c]{0.22\linewidth}
        \centering
        \scriptsize FaceLift (PSGS)
    \end{minipage}%
    \hspace{0.01\linewidth}
    \begin{minipage}[c]{0.76\linewidth}
        \centering
        \includegraphics[width=1\linewidth]{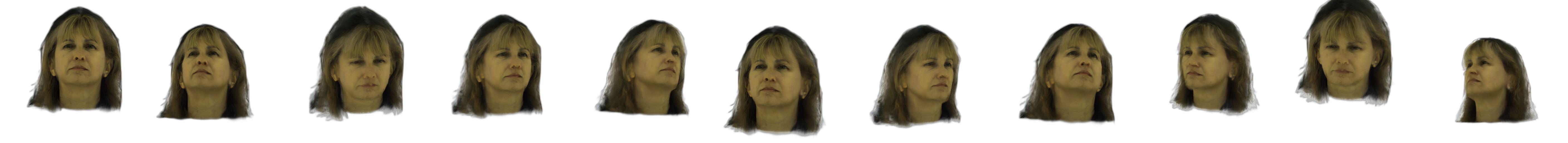}
    \end{minipage}
    \par\vspace{1mm}
    \begin{minipage}[c]{0.22\linewidth}
        \centering
        \scriptsize \textit{Ours +~\cite{psgs}}
    \end{minipage}%
    \hspace{0.01\linewidth}
    \begin{minipage}[c]{0.76\linewidth}
        \centering
        \includegraphics[width=1\linewidth]{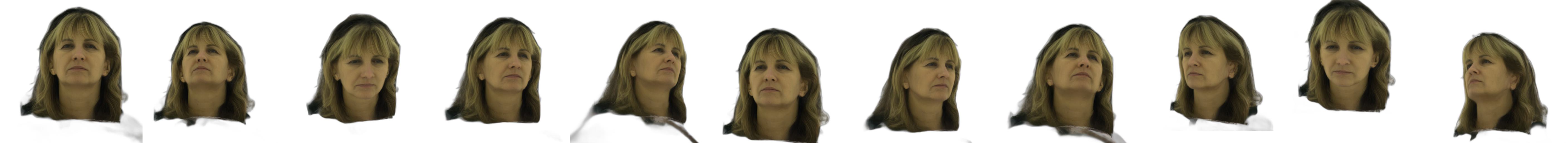}
    \end{minipage}
    \par\vspace{1mm}
    \begin{minipage}[c]{0.22\linewidth}
        \centering
        \scriptsize \textit{Ours MO +~\cite{psgs}}
    \end{minipage}%
    \hspace{0.01\linewidth}
    \begin{minipage}[c]{0.76\linewidth}
        \centering
        \includegraphics[width=1\linewidth]{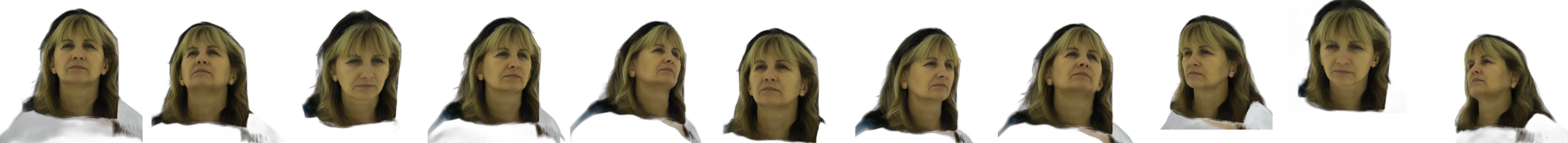}
    \end{minipage}
    \par\vspace{1mm}
    \begin{minipage}[c]{0.22\linewidth}
        \centering
        \scriptsize \textit{Ours MI-MO +~\cite{psgs}}
    \end{minipage}%
    \hspace{0.01\linewidth}
    \begin{minipage}[c]{0.76\linewidth}
        \centering
        \includegraphics[width=1\linewidth]{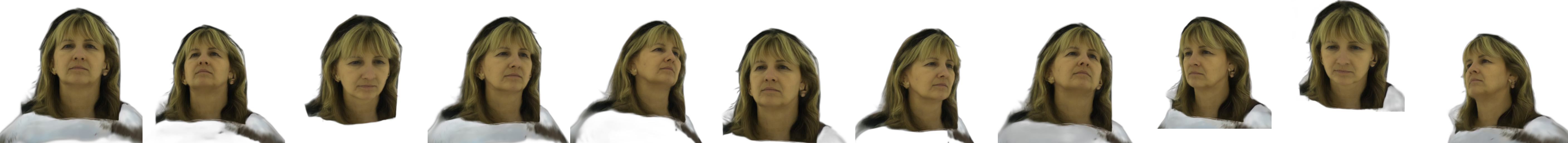}
    \end{minipage}
    \caption{Qualitative results on AVA-256~\cite{ava256} for subject 20210901--0833--LAS440}
    \label{fig:ava1}
\end{figure*}

\subsection{Ablation studies}

\begin{figure*}[ht]
    \centering
    \begin{minipage}[t]{0.48\textwidth}
        \centering
        \begin{minipage}[c]{0.25\linewidth}
            \centering
            \tiny AV~\cite{archonview} + $2^{nd}$ stage ($K=12$)
        \end{minipage}%
        \hspace{0.01\linewidth}
        \begin{minipage}[c]{0.7\linewidth}
            \centering
            \includegraphics[width=1\linewidth]{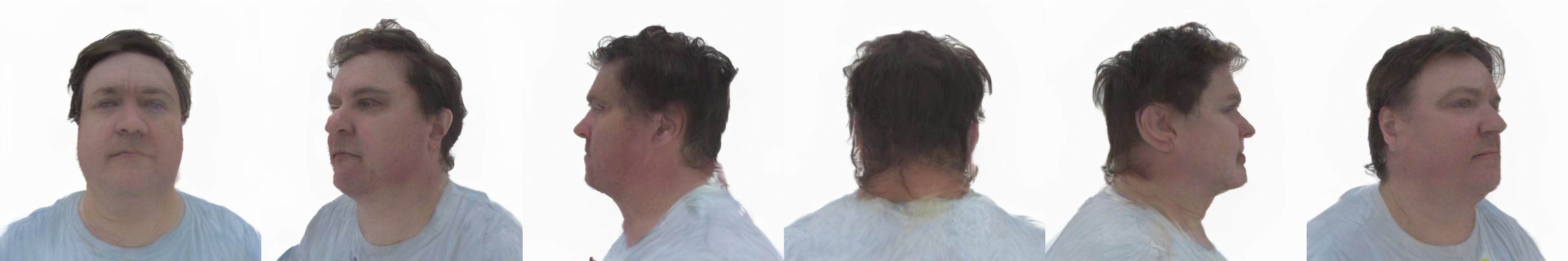}
        \end{minipage}
        \par\vspace{1mm}

        \begin{minipage}[c]{0.25\linewidth}
            \centering
            \tiny AV~\cite{archonview} + $2^{nd}$ stage ($K=10$)
        \end{minipage}%
        \hspace{0.01\linewidth}
        \begin{minipage}[c]{0.7\linewidth}
            \centering
            \includegraphics[width=1\linewidth]{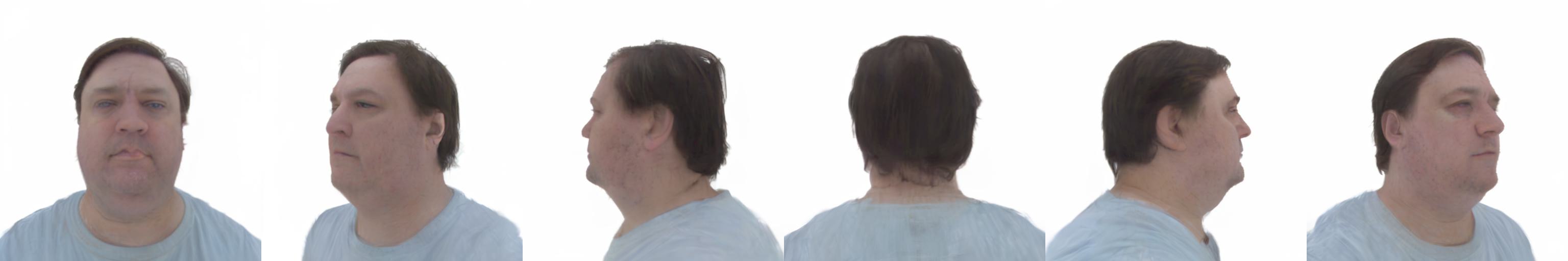}
        \end{minipage}
        \par\vspace{1mm}


        \begin{minipage}[c]{0.25\linewidth}
            \centering
            \tiny AV~\cite{archonview} + $1^{st}$+$2^{nd}$ stage ($K=10$)
        \end{minipage}%
        \hspace{0.01\linewidth}
        \begin{minipage}[c]{0.7\linewidth}
            \centering
            \includegraphics[width=1\linewidth]{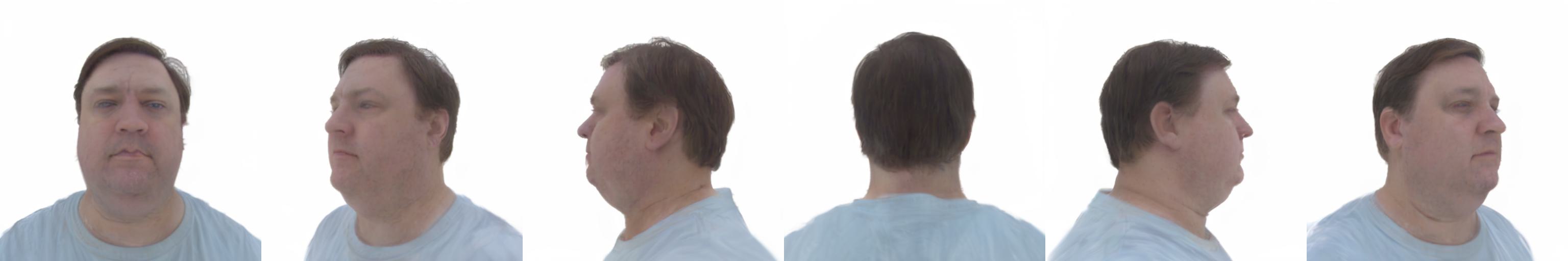}
        \end{minipage}
        \par\vspace{1mm}

        \begin{minipage}[c]{0.25\linewidth}
            \centering
            \tiny VAR~\cite{VAR} + $2^{nd}$ stage
        \end{minipage}%
        \hspace{0.01\linewidth}
        \begin{minipage}[c]{0.7\linewidth}
            \centering
            \includegraphics[width=1\linewidth]{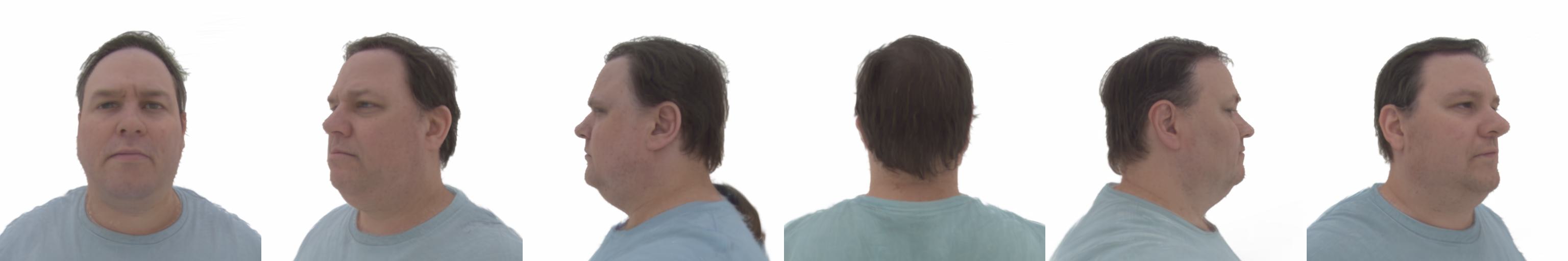}
        \end{minipage}

        \begin{minipage}[c]{0.25\linewidth}
            \centering
            \tiny GT
        \end{minipage}%
        \hspace{0.01\linewidth}
        \begin{minipage}[c]{0.7\linewidth}
            \centering
            \includegraphics[width=1\linewidth]{figs/nvs1/gt.jpg}
        \end{minipage}
    \end{minipage}
    \hfill
    \begin{minipage}[t]{0.48\textwidth}
        \vspace{0.1cm}
        \centering
        \begin{minipage}[c]{0.25\linewidth}
            \centering
            \tiny CFG=0
        \end{minipage}%
        \hspace{0.01\linewidth}
        \begin{minipage}[c]{0.7\linewidth}
            \centering
            \includegraphics[width=1\linewidth]{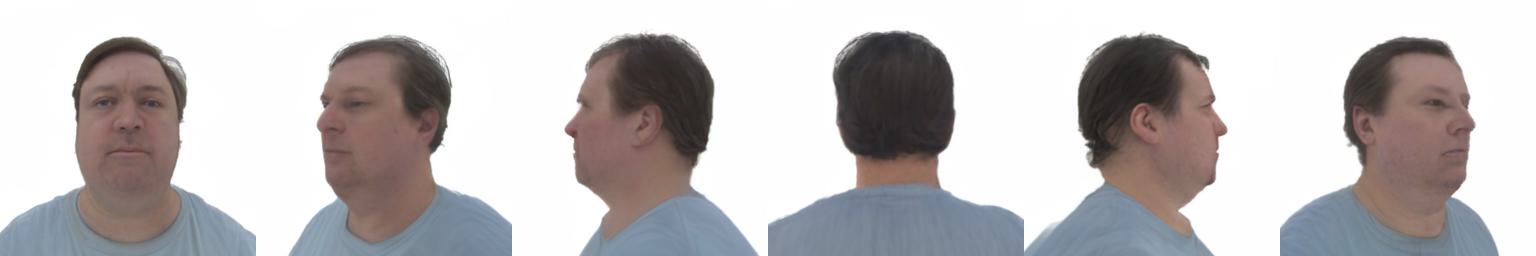}
        \end{minipage}
        \par\vspace{1mm}

        \begin{minipage}[c]{0.25\linewidth}
            \centering
            \tiny CFG=1
        \end{minipage}%
        \hspace{0.01\linewidth}
        \begin{minipage}[c]{0.7\linewidth}
            \centering
            \includegraphics[width=1\linewidth]{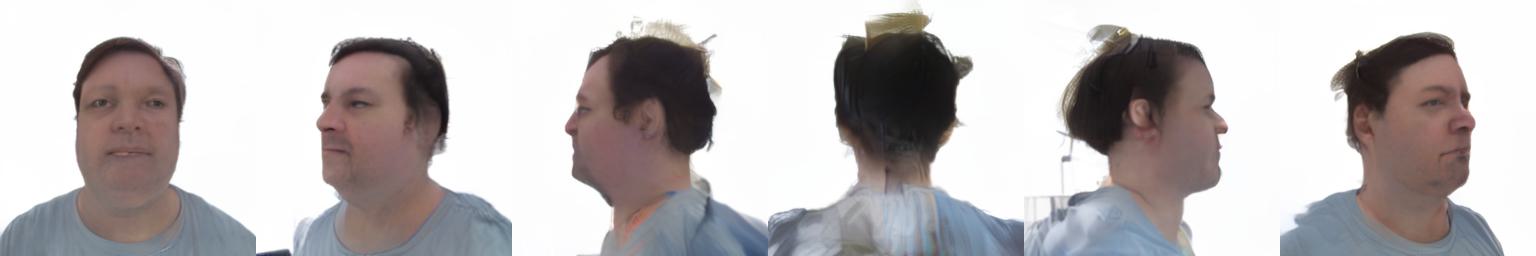}
        \end{minipage}
        \par\vspace{1mm}

        \begin{minipage}[c]{0.25\linewidth}
            \centering
            \tiny CFG=3
        \end{minipage}%
        \hspace{0.01\linewidth}
        \begin{minipage}[c]{0.7\linewidth}
            \centering
            \includegraphics[width=1\linewidth]{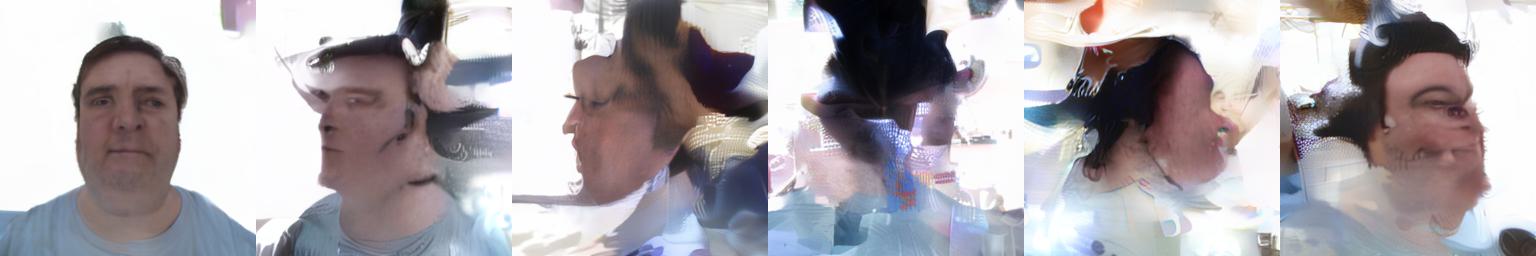}
        \end{minipage}
        \par\vspace{1mm}



        \begin{minipage}[c]{0.25\linewidth}
            \centering
            \tiny GT
        \end{minipage}%
        \hspace{0.01\linewidth}
        \begin{minipage}[c]{0.7\linewidth}
            \centering
            \includegraphics[width=1\linewidth]{figs/nvs1/gt.jpg}
        \end{minipage}
    \end{minipage}
    \caption{(Left) Adapting the model to single-output 512$\times$512 resolution. The first three rows use ArchonView~\cite{archonview} (d30) as the initial checkpoint, which was originally trained for NVS of objects at 256$\times$256, while the fourth row uses VAR~\cite{VAR} (d36), which was originally trained for image generation at 512$\times$512. VQ-VAE is not fine-tuned in any of the models. (Right) Impact of Classifier-Free Guidance (CFG) on our d12 multi-output model during inference. Similar effects are noticed on bigger models.}
    \label{fig:adapting_and_cfg}
\end{figure*}

\begin{table*}[h!]
\centering
\caption{Ablation study of the NVS-stage of our pipeline on PSGS~\cite{psgs} validation set (6 canonical views) for single-input single-output models. K=10 unless otherwise specified, and using around half the training steps of the final model.}
\label{tab:ablation}
\scriptsize
\resizebox{\textwidth}{!}{
\begin{tabularx}{\textwidth}{*{6}{>{\centering\arraybackslash}X}|*{2}{>{\centering\arraybackslash}X}}
Depth & Resolution & Checkpoint & $1^{st}$ stage & $2^{nd}$ stage & VQ-VAE & PSNR $\uparrow$ & ArcFace $\uparrow$ \\
\hline
d12 & 256 & AV~\cite{archonview} & & & default & 12.02 & 17.01 \\
d12 & 256 & AV~\cite{archonview} & & \checkmark & default & 19.77 & 39.20 \\
d12 & 256 & Scratch & & \checkmark & default & 19.68 & 34.87 \\
d36 & 512 & VAR~\cite{VAR} & & \checkmark & default & 16.72 & 38.17 \\
d12 & 512 K=12 & AV~\cite{archonview} & & \checkmark & default & 15.70 & 33.71 \\
d12 & 512 & AV~\cite{archonview} & & \checkmark & default & 21.43 & 57.75 \\
d30 & 512 & AV~\cite{archonview} & & \checkmark & default & 20.34 & 54.51 \\
d30 & 512 & AV~\cite{archonview} & \checkmark & \checkmark & default & 21.66 & 63.38 \\
d30 & 512 & AV~\cite{archonview} & \checkmark & \checkmark & fine-tuned & 22.24 & 67.72 \\
\end{tabularx}
}
\vspace{-0.5cm}
\end{table*}

In order to better understand the impact of the different components of our method, we perform an ablation study on the validation set of PSGS~\cite{psgs} with the previously defined 6 canonical views. In~\cref{tab:ablation} we see that the original resolution of ArchonView~\cite{archonview}, 256$\times$256, is not sufficient to achieve good results on faces. Moreover, training the model from scratch on face data does not lead to good results, as well as starting from a powerful non-NVS model such as VAR-d36~\cite{VAR}. Even though such model is already trained on 512x512 images, it produces high-quality results but without correctly preserving identity and view consistency, see \cref{fig:adapting_and_cfg} (left) row 4. Thus, adapting the existing ArchonView 256x256 model to our target resolution is crucial, and has to be approached by keeping the number of scales the same, as conjectured in \cref{sec:training_strategy}. In \cref{tab:ablation} row 5-6 and \cref{fig:adapting_and_cfg} (left) row 1-2 we see that this choice is key to achieve good results. Moreover, while small models (d12) can be easily adapted to the new resolution, larger models (d30) struggle to converge, justifying the need for our first training stage, which allows the model to adapt to the new resolution before being trained on the target data. Finally, the fine-tuned VQ-VAE slightly improves the results.\\
In \cref{fig:adapting_and_cfg} (right) we also show the negative effect of inference-time classifier-free guidance during the third training stage. Moreover, compared to the d12 256$\times$256 CFG$=$0 setting seen in the figure, which achieves a PSNR of 19.99 and an ArcFace of 0.4044, the same model without classifier-free guidance during the third training stage achieves a PSNR of 20.28 and an ArcFace of 0.4296, corroborating the importance of this choice.

\subsection{Limitations}

While achieving good results across all the metrics and experiments, our method still has some limitations. In particular, the training data only contains neutral expressions and no accessories, thus the model struggles to correctly generalize to unseen expressions and accessories, such as hats. These limitations are similar to the ones reported by FaceLift~\cite{facelift}. Additionally, we noticed that all the methods tend to suffer from background color bleeding. This is noticeable in \cref{fig:ava1}, where the black background is not perfectly removed and a portion of the hair is incorrectly tinted in black in the other views. Brighter background colors worsen the issue. More fine-grained background removal techniques would surely help to mitigate this, but it is still a common problem in the field. 

  \section{Conclusions}
We have presented a novel method for photorealistic view-consistent novel view synthesis of human faces using next-scale transformers, which can be applied to other domains as well, such as full bodies. Our method achieves state-of-the-art results on the PSGS dataset, outperforming previous methods in terms of both quantitative metrics and qualitative visual quality, and is also effective on the Ava-256 dataset, showing that it can generalize well to different datasets and scenarios. We believe that future work can focus on addressing limitations by exploring more diverse training data, as well as recent developments in the next-scale VAR framework, and further specializing the architecture to different scenarios such as full body reconstruction.

\fi
{
    %
    %
    \small
    \bibliographystyle{splncs04}
    \bibliography{main}
}

\clearpage
\renewcommand\thefigure{A.\arabic{figure}} 
  \renewcommand\thetable{A.\arabic{table}} 
  \renewcommand\thesection{A.\arabic{section}}
  \setcounter{figure}{0}
  \setcounter{table}{0}
  \setcounter{section}{0}
  
  \clearpage



\end{document}